\documentclass{article}
\usepackage[T1]{fontenc}
\usepackage[utf8]{inputenc}
\usepackage{preprint,times}
\ppfinalcopy
\usepackage{amsmath,amssymb,amsthm,booktabs,array,longtable,tabularx,multirow}
\usepackage{enumitem}
\usepackage{graphicx,xcolor,float,placeins,pdflscape}
\usepackage{colortbl}

\usepackage{tikz}
\usetikzlibrary{arrows.meta,positioning,fit,calc}
\usepackage{xurl}
\usepackage[hidelinks]{hyperref}
\usepackage{microtype}

\newcommand{\bench}{WristHARBench}

\newcommand{\R}{\mathbb{R}}
\newcommand{\tr}{\mathrm{tr}}
\newcommand{\va}{\mathrm{va}}
\newcommand{\te}{\mathrm{te}}

\newcolumntype{L}[1]{>{\raggedright\arraybackslash}p{#1}}

\title{WISP: Composable Representations for\\ Wrist-Worn Activity Recognition}
\author{\begin{tabular}[t]{@{}>{\normalfont}p{0.47\textwidth}@{\hspace{0.03\textwidth}}>{\normalfont}p{0.47\textwidth}@{}}\textbf{Zipeng Wu} & \textbf{Shiqiao Zhou} \\ School of Mathematics & School of Computer Science \\ University of Birmingham, UK & University of Birmingham, UK \\ \texttt{zxw365@student.bham.ac.uk} & \texttt{sxz363@student.bham.ac.uk} \\[12pt] \textbf{Fabian Spill} & \textbf{James Andrews} \\ School of Mathematics & School of Mathematics \\ University of Birmingham, UK & University of Birmingham, UK \\ \texttt{f.spill@bham.ac.uk} & \texttt{j.w.andrews@bham.ac.uk}\end{tabular}}
\hypersetup{pdftitle={WISP: Composable Representations for Wrist-Worn Activity Recognition},pdfauthor={}}
\begin{document}
\hypersetup{pdfauthor={Zipeng Wu, Shiqiao Zhou, Fabian Spill, James Andrews}}
\maketitle
\lhead{Preprint}
\begin{abstract}
Smartwatches and fitness bands are among the most common wearable devices, and large national health studies record movement at the wrist. Yet existing human activity recognition benchmarks mix sensors worn on different parts of the body, so they do not show how well activities can be recognised from the wrist alone. We introduce \bench{}, a benchmark that standardises participant-disjoint evaluation across nine wrist-acceleration datasets and two additional multichannel inertial datasets. We hypothesise that different human activities leave complementary patterns in wrist motion, such as local shape, rhythm, temporal order and cross-axis relationships, and propose WISP (Wrist-Informed Space of Predictive Representations), a composable representation space that combines representations of these patterns with compatible classifiers. Across the nine primary datasets, WISP with random search achieves \textbf{70.20\% mean Macro-F1, compared with 59.62\%} for the strongest of 18 external baselines. Controlled ablations and baseline comparisons indicate that \textbf{recognition improves mainly when a representation adds motion patterns that the others lack, rather than when it represents the same pattern in a different way}.
Code and saved results are available at \url{https://huggingface.co/Zipeng365/WISP}, and the benchmark data at \url{https://huggingface.co/datasets/Zipeng365/WristHARBench}.
\end{abstract}

\section{Introduction}
\label{sec:intro}
Human activity recognition (HAR) from wearable sensors supports health monitoring and assisted living \citep{lara2013survey,bulling2014tutorial,chen2021deep}. Smartwatches and wristbands are among the most common consumer wearables, and they record wrist movement during daily activities such as walking, eating and cleaning \citep{henriksen2018fitness,huhn2022impact}. National health studies also collect movement data at the wrist or with consumer wearables: UK Biobank recorded wrist accelerometry from over 100,000 participants \citep{doherty2017ukb}, NHANES from 14,705 US participants \citep{belcher2021nhanes}, and the All of Us Research Program has collected Fitbit data from more than 59,000 participants \citep{patten2026allofus}. HAR from the wrist is therefore the setting in which activity recognition is most likely to be used at population scale.

However, existing multi-dataset HAR benchmarks combine sensors worn at the wrist and at other body locations \citep{burzer2025whardatasets,burzer2026whararena}, \textbf{so their results do not show how well a method recognises activities from the wrist alone}. A wrist-only benchmark matches where large-scale data are available and where deployment costs least, since a single wrist-worn device needs no sensors elsewhere on the body. Evaluation protocols also affect reported performance: overlapping windows combined with participant-dependent evaluation can inflate scores \citep{dehghani2019windows}, and even wrist-specific studies have used different participant-split protocols across datasets \citep{yuan2024ssl}. We therefore introduce \bench{}, whose primary comparison uses the wrist as the only sensing location: nine wrist-acceleration datasets under repeated participant-disjoint splits, with two multichannel inertial tasks as additional analyses.

\begin{figure}[t]
\centering
\includegraphics[width=\linewidth]{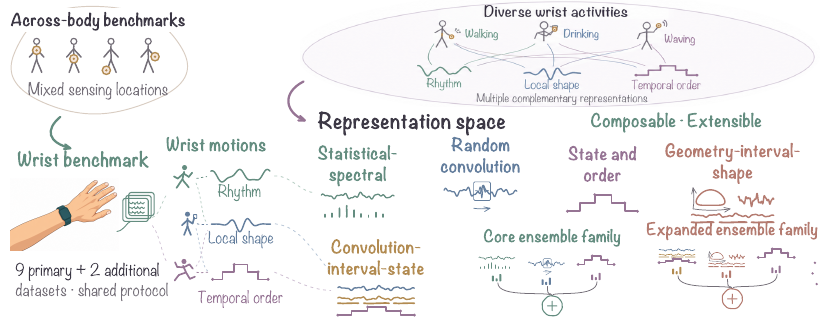}
\caption{Motivation and overview of WristHARBench and WISP.
WristHARBench fixes the sensing location and participant-disjoint protocol across nine primary and two additional datasets.
Different wrist activities exhibit complementary motion patterns, motivating a composable, extensible representation space. WISP organises this space into five representation families and two ensemble families.
Connections illustrate multiple representations of an activity and patterns shared across activities.
Ensemble families combine separately fitted, class-aligned scores. Signals and activity patterns are schematic.}
\label{fig:wisp_overview}
\end{figure}

Wrist HAR is a multichannel time-series classification problem: each window is a short three-axis acceleration signal \citep{ruiz2021multivariate}. Most recent HAR methods are neural, including supervised architectures and self-supervised pretrained encoders \citep{abedin2021attend,yuan2024ssl}. Time-series classification has developed a complementary set of representations, each designed for a different temporal structure: random convolutions for local shape at several time scales, interval statistics for value distributions within segments, and shapelets for discriminative subsequences \citep{dempster2021minirocket,tan2022multirocket,dempster2023hydra,middlehurst2020cif,dempster2024quant,guillaume2022rdst}. \textbf{Which temporal structures matter for wrist activities, and whether combining them helps, has not been tested under a common protocol.}

A wrist recording can superpose motion from whole-body locomotion, arm movement and local interactions with objects. Prior work shows that inertial signals are sensitive to sensor location and device heterogeneity \citep{bulling2014tutorial,atallah2011positioning,stisen2015hhar}; wrist motion sources may leave patterns at different time scales and across sensor axes. Together, these observations suggest a testable hypothesis: wrist HAR may benefit from representations that preserve and combine complementary motion patterns, such as local shape, rhythm, temporal order and cross-axis relationships. \textbf{We test this hypothesis by varying representation components while holding the classifier and evaluation protocol fixed.}

We propose \emph{WISP} (Wrist-Informed Space of Predictive Representations), which \textbf{combines wrist-signal representations with compatible classifiers and optional temporal decoding}. Within WISP, we design five representation families and two score ensembles. Among them, WISP-CIS (convolution--interval--state) combines a random convolution representation, an interval representation and a spectral, temporal and cross-axis representation before fitting a shared classifier. Figure~\ref{fig:wisp_overview} connects the benchmark and representation-space motivations to the seven families.

Our contributions are:
\begin{enumerate}
\setlength{\itemsep}{1pt}\setlength{\parsep}{0pt}
\item \textbf{WristHARBench.} We provide a common participant-disjoint protocol for nine primary wrist-acceleration datasets and two additional multichannel inertial tasks.\footnote{Benchmark data: \url{https://huggingface.co/datasets/Zipeng365/WristHARBench}}
\item \textbf{WISP.} We design five representation families and two score ensembles within a composable space. WISP-CIS combines random convolution, interval, and spectral, temporal and cross-axis representations before a shared classifier.\footnote{Code, saved results and model checkpoints: \url{https://huggingface.co/Zipeng365/WISP}}
\item \textbf{Empirical findings.} Fixed WISP-CIS achieves 66.14\% mean Macro-F1 without recipe search. Controlled ablations support combining random convolution with the spectral, temporal and cross-axis representation. Matched-budget comparisons identify structured random search as a strong strategy within WISP, with no consistent gain from evolutionary adaptation.
\end{enumerate}

\section{Related Work}
\label{sec:related}
DeepConvLSTM combines convolutional and recurrent layers for activity recognition \citep{ordonez2016deep}. Attend-and-Discriminate models relationships between sensor channels and learns discriminative representations \citep{abedin2021attend}. HARNet learns wrist-accelerometry representations through large-scale self-supervised pretraining \citep{yuan2024ssl}.

WHAR Datasets standardises data handling, and WHAR Arena compares wearable HAR models under a shared cross-subject protocol \citep{burzer2025whardatasets,burzer2026whararena}. These resources include recordings from smartphones and wearable sensors at multiple body locations. HARNet evaluates eight wrist datasets using leave-one-participant-out evaluation for smaller datasets and fivefold participant-disjoint evaluation for the others \citep{yuan2024ssl}. \bench{} focuses on wrist recordings and applies the same repeated participant-disjoint evaluation protocol across its datasets.

\looseness=-1 MiniRocket and MultiRocket use convolutional kernels and pooling, while RDST uses dilated shapelets \citep{dempster2021minirocket,tan2022multirocket,guillaume2022rdst}. MR-Hydra combines MultiRocket and Hydra representations before fitting a classifier \citep{dempster2023hydra}. HARNAS searches neural architectures, while AutoAugHAR searches data augmentation policies for sensor signals \citep{wang2021harnas,zhou2024autoaughar}. WISP searches over predefined wrist-signal representation families, compatible classifiers and optional temporal decoding.

\section{WristHARBench and WISP}
\label{sec:method}

\subsection{Problem formulation}
\label{sec:problem}
We formulate wrist HAR as supervised classification of multichannel signal windows into $K$ activity classes \citep{ruiz2021multivariate}. A dataset of $N_w$ windows is written as
\begin{equation}
 \mathcal D=\{(X_i,y_i,u_i,o_i)\}_{i=1}^{N_w},\qquad
 X_i\in\R^{T\times C},\quad y_i\in\mathcal Y=\{0,\ldots,K-1\}.
\label{eq:problem}
\end{equation}
Here $X_i$ contains $T$ time samples from $C$ sensor channels, and $y_i$ is its activity label. The participant identifier $u_i$ determines data splitting, and $o_i$ records the window's order within that participant for optional temporal decoding.

A classifier with fitted parameters $\omega$ produces normalised class scores, and direct prediction selects the highest-scoring class:
\begin{equation}
 f_\omega:\R^{T\times C}\rightarrow\Delta^{K-1},\qquad
 \mathbf s_i=f_\omega(X_i),\qquad
 \widehat y_i=\arg\max_{k\in\mathcal Y}[\mathbf s_i]_k,
\label{eq:classifier}
\end{equation}
where $\Delta^{K-1}=\{\mathbf s\in\R^K:s_k\geq0,\sum_{k=0}^{K-1}s_k=1\}$. These scores need not be calibrated probabilities. We partition $\mathcal D$ into training, validation and test sets, denoted by $\mathcal D_\tr$, $\mathcal D_\va$ and $\mathcal D_\te$. Let $\mathcal U_a$ be the set of participant identifiers in $\mathcal D_a$, for $a\in\{\tr,\va,\te\}$. Participant-disjoint evaluation requires
\begin{equation}
 \mathcal U_\tr\cap\mathcal U_\va=
 \mathcal U_\tr\cap\mathcal U_\te=
 \mathcal U_\va\cap\mathcal U_\te=\varnothing.
\label{eq:participant_split}
\end{equation}

Within each split, candidate models are fitted on $\mathcal D_\tr$, with configurations fixed in advance or selected using $\mathcal D_\va$. Each predefined or validation-selected configuration is refitted from scratch on $\mathcal D_\tr\cup\mathcal D_\va$ for test evaluation. For temporal decoding, windows are grouped by $u_i$ and ordered by $o_i$.

\subsection{WristHARBench}
\label{sec:benchmark}
\label{sec:datasets}
The primary comparison in \bench{} uses nine datasets of three-axis wrist acceleration (Table~\ref{tab:datasets}). HARMES is evaluated separately for each wrist and counted as one dataset in aggregate comparisons. PAMAP2 (hand) and MHEALTH (right lower arm) provide two additional nine-channel inertial datasets for sensitivity analysis \citep{pamap2Dataset,mhealthDataset}. Appendix~\ref{app:export_rules} specifies the windowing, label processing and source mapping.
\begin{table}[htb]
\centering\footnotesize\setlength{\tabcolsep}{4pt}
\caption{\bench{} datasets and processed inputs. $N_u$: participants; $N_w$: retained windows; $C$ is the exported channel count before baseline-specific adapters; $K$ follows Section~\ref{sec:problem}. The first nine are primary tasks; HARMES has 44,323 windows per wrist, is evaluated separately and paired for reporting; PAMAP2 (hand) and MHEALTH (right lower arm) are additional nine-channel tasks.}
\label{tab:datasets}
\begin{tabular*}{\linewidth}{@{\extracolsep{\fill}}lrrrrrr@{}}
\toprule Dataset & $N_u$ & $N_w$ & Hz & Window (s) & $C$ & $K$\\\midrule
CAPTURE-24 \citep{chan2024capture} &151&50,459&25&10&3&4\\
WISDM-watch \citep{weiss2019wisdm} &51&74,499&20&5&3&18\\
ADL-Wrist \citep{adlWristDataset} &16&4,342&32&5&3&14\\
IUWDS \citep{karas2021iuwds} &32&35,442&100&5&3&6\\
PAAL-ADL \citep{climent2022paal} &52&14,384&32&5&3&24\\
HANDY \citep{acici2018handy} &30&7,163&52&5&3&10\\
DOMINO \citep{arrotta2023domino} &25&11,627&100&5&3&14\\
HARMES, paired \citep{harmesDataset2026} &20&44,323&50&5&3&16\\
GOTOV \citep{paraschiakos2020gotov} &35&33,801&83&5&3&16\\\midrule
PAMAP2 \citep{pamap2Dataset} &9&7,637&100&5&9&12\\
MHEALTH \citep{mhealthDataset} &10&2,561&50&5&9&12\\
\bottomrule
\end{tabular*}
\end{table}

\label{sec:evaluation}
\label{sec:metrics}
We use five repeated participant-disjoint splits, assigning approximately 70\%, 10\% and 20\% of participants to training, validation and test, respectively. All methods use the same windows and participant splits, with baseline-specific input preprocessing described in Appendix~\ref{app:baseline_settings}. Test sets may overlap across repeats. Macro-F1, the primary metric, averages over classes present in true or predicted test labels; worst-class F1 uses the full dataset vocabulary. We also report accuracy and balanced accuracy (Appendix~\ref{app:metric_definitions}). For each split, we average scores over model seeds within each wrist, then average the two HARMES wrists. Each dataset is summarised by the mean and standard deviation across the five splits. Aggregate scores give equal weight to the nine primary datasets. We report mean paired differences between methods across the nine primary datasets, dataset-bootstrap confidence intervals and Holm-adjusted exact two-sided sign-flip tests.

\subsection{Wrist-Informed Space of Predictive Representations}
\label{sec:representation}\label{sec:families}\label{sec:selection}
WISP includes five representation families and two score ensembles (Table~\ref{tab:families}). Each representation family produces a fixed-length representation that a fitted classifier maps to $K$ class scores. The two ensembles combine scores from independently fitted classifiers after aligning them to the same activity classes.

\begin{table}[htb]
\centering\footnotesize\setlength{\tabcolsep}{3.5pt}\renewcommand{\arraystretch}{0.96}
\caption{WISP methods: five representation families and two score ensembles.}
\label{tab:families}
\begin{tabularx}{\linewidth}{@{}L{.30\linewidth}X@{}}
\toprule Method & Computation\\\midrule
WISP-SS (statistical--spectral) & Window and segment statistics, spectra, autocorrelations, quantile-state summaries and channel correlations $\rightarrow$ tree-based or logistic classifier.\\
WISP-RC (random convolution) & Pooled responses of random multichannel dilated kernels $\rightarrow$ Ridge or logistic classifier.\\
WISP-SO (state and order) & Quantile-state summaries, window and segment statistics, and channel correlations $\rightarrow$ ExtraTrees.\\
WISP-GIS (geometry--interval--shape) & Geometry, interval statistics, subsequence matches, and channel and magnitude summaries $\rightarrow$ tree-based or logistic classifier.\\
WISP-CIS (convolution--interval--state) & Concatenated random convolution, interval, spectral, temporal and cross-axis representations $\rightarrow$ Ridge or logistic classifier.\\\midrule
WISP-CSE (core score ensemble) & Equal-weight average of scores from WISP-SS, reduced WISP-RC and WISP-SO.\\
WISP-ESE (expanded score ensemble) & Weighted average of scores from reduced WISP-CIS, reduced WISP-GIS and WISP-SO.\\
\bottomrule
\end{tabularx}
\par\smallskip
\raggedright Reduced ensemble members use fewer kernels, intervals or sampled subsequences (Appendix~\ref{app:family_settings}).
\end{table}

For an input window $X$, we omit the window index $i$ and standardise each channel over time:
\begin{equation}
 \widetilde X_{t,c}=\frac{X_{t,c}-\mu_c(X)}{\bar\sigma_c(X)},
 \qquad t=1,\ldots,T,\quad c=1,\ldots,C,
\label{eq:window_standardisation}
\end{equation}
Here $\mu_c(X)$ is the channel mean and $\bar\sigma_c(X)$ is its population standard deviation, replaced by 1 if below $10^{-6}$.

\begin{enumerate}[label=(\arabic*),leftmargin=1.8em,labelsep=.35em,itemsep=2pt,parsep=0pt,topsep=2pt]
\item WISP-SS (statistical--spectral) summarises the standardised channels and their first differences. It also uses the Euclidean norm of the first $\min(3,C)$ channels in each view. It computes whole-window and segment statistics, spectral summaries and autocorrelations from these views (Appendix~\ref{app:operators}). It also records the occupancies, transitions and transition entropy of window-relative quantile states, together with correlations between the standardised channels. The resulting values are concatenated and classified by a tree-based or logistic model; the fixed configuration uses ExtraTrees.

\item WISP-RC (random convolution) uses randomly sampled kernels to represent local waveforms at multiple time scales \citep{dempster2020rocket,tan2022multirocket}. Each kernel $j$ has a sampled length $L_j$, dilation $d_j$, channel subset $\mathcal C_j$, weights $k_{j,c,\ell}$ and bias $b_j$; weights are centred and normalised (Appendix~\ref{app:sampled}). For $t=1,\ldots,T-d_j(L_j-1)$, its response is
\begin{equation}
 r_j(t)=b_j+\sum_{c\in\mathcal C_j}\sum_{\ell=0}^{L_j-1}
 k_{j,c,\ell}\widetilde X_{t+d_j\ell,c}.
\label{eq:random_convolution}
\end{equation}
For each kernel, we pool its response over time using the proportion of positive values, maximum, mean and population standard deviation. Concatenating the pooled values from $K_R$ kernels gives a $4K_R$-dimensional representation. A Ridge or logistic classifier uses this representation; the fixed configuration uses 384 kernels and Ridge.

\item WISP-SO (state and order) uses a symbolic representation of each selected scalar signal $v=(v_1,\ldots,v_T)$, with states defined by within-window quantiles \citep{lin2007sax}. Let $B_q$ be the number of states and $\tau_b(v)$ the empirical $b/B_q$ quantile, for $b=1,\ldots,B_q-1$. The state index $a_t$, occupancy $p_a$ and row-normalised transition matrix $Q$ are
\begin{equation}
 \begin{aligned}
 a_t&=\sum_{b=1}^{B_q-1}\mathbf 1[v_t>\tau_b(v)],
 &p_a&=\frac{1}{T}\sum_{t=1}^{T}\mathbf 1[a_t=a],\\
 Q_{ab}&=\frac{\sum_{t=1}^{T-1}\mathbf 1[a_t=a,\,a_{t+1}=b]}
 {\max\!\left(1,\sum_{t=1}^{T-1}\mathbf 1[a_t=a]\right)}.
 \end{aligned}
\label{eq:state_order}
\end{equation}
Here $a,b\in\{0,\ldots,B_q-1\}$; rows with no observed transitions remain zero. We concatenate the occupancies $p_a$, entries of $Q$, the sum of its row entropies, whole-window and segment statistics, and channel correlations. The fixed configuration uses $B_q=8$ and ExtraTrees with 160 trees; signal views and segment settings are specified in Appendix~\ref{app:family_settings}.

\item WISP-GIS (geometry--interval--shape) combines cross-axis geometry, interval statistics and subsequence matches with channel and magnitude summaries. The geometry part records the eigenvalues of the covariance of the first $\min(C,3)$ standardised channels. The interval part computes per-channel statistics and quantiles over fixed and sampled intervals \citep{middlehurst2020cif,dempster2024quant}. The shape part records the minimum root-mean-square distance and best-match position of subsequences sampled from training windows \citep{guillaume2022rdst}. A tree-based or logistic classifier uses the concatenated representation; the fixed configuration uses 24 sampled intervals, 32 subsequences and ExtraTrees (Appendix~\ref{app:operators}).

\item WISP-CIS (convolution--interval--state) concatenates three representations for a single classifier:
\begin{equation}
 \mathbf z_{RIS}(X)=\mathbf z_R(\widetilde X)\Vert\mathbf z_I(\widetilde X)\Vert\mathbf z_S(\widetilde X),
 \qquad \mathbf s(X)=h_\omega\!\left(\mathbf z_{RIS}(X)\right).
\label{eq:cis}
\end{equation}
Here $\Vert$ denotes concatenation; $\mathbf z_R$ contains the four pooled responses per kernel from WISP-RC, and $\mathbf z_I$ contains the interval statistics used in WISP-GIS. The spectral, temporal and cross-axis representation $\mathbf z_S$ contains spectral summaries, autocorrelations, state occupancies, transitions and transition entropy with $B_q=6$, and cross-axis correlations. The mapping $h_\omega$ standardises the concatenated representation using training-set statistics, applies a class-balanced Ridge or logistic classifier, and converts its outputs to the class scores in Equation~\ref{eq:classifier}. The fixed configuration uses 384 kernels, 24 sampled intervals and Ridge. Controlled ablations in Section~\ref{sec:ingredient_results} vary the included representations while keeping the classifier, bank sizes and evaluation protocol fixed.

\item WISP-CSE (core score ensemble) fits WISP-SS, reduced WISP-RC and WISP-SO independently. It averages their score vectors with equal weights after aligning them to the same $K$ activity classes in the same order.

\item WISP-ESE (expanded score ensemble) fits reduced WISP-CIS and WISP-GIS models and a WISP-SO model independently. It combines their class-aligned score vectors using weights $w=(w_1,w_2,w_3)$:
\begin{equation}
 \mathbf s_{\mathrm{ESE}}(X)=w_1\mathbf s_{\mathrm{CIS}}(X)+w_2\mathbf s_{\mathrm{GIS}}(X)+w_3\mathbf s_{\mathrm{SO}}(X),
 \qquad w_j\geq0,\quad\sum_{j=1}^{3}w_j=1.
\label{eq:expanded_ensemble}
\end{equation}
The fixed weights and member configurations are specified in Appendix~\ref{app:family_settings}; search may vary $w$. For both ensembles, optional temporal decoding is applied once after score averaging.
\end{enumerate}

Each method either predicts the highest-scoring class per window or applies offline Viterbi decoding $H$ to each participant's full sequence, which uses later windows (Appendix~\ref{app:heads}). We evaluate one predefined recipe for each of the seven methods, and all fixed recipes use $H$. WISP-Select5 chooses among the five fixed representation recipes by validation Macro-F1, and WISP-Select7 also includes the two score ensembles; neither selector changes the decision rule. WISP-Random and WISP-Evolution vary the method and its allowed settings, including $H$, from the same nine initial recipes with 329 candidate evaluations per run. Both return the candidate with the highest validation utility $U$, which combines Macro-F1, worst-class F1 and performance under four signal perturbations with penalties for run time and complexity. WISP-Random applies random changes to freshly seeded templates without using earlier scores, whereas WISP-Evolution selects parents from evaluated candidates and updates mutation preferences from their utilities (Appendix~\ref{app:search}). All configurations follow the refitting protocol in Section~\ref{sec:problem}.

\section{Experimental Study}
\label{sec:experiments}

\label{sec:experimental_setup}
\label{sec:experiment_matrix}
To evaluate WISP's effectiveness, we compare WISP-Random and WISP-Evolution with eighteen external baselines on WristHARBench. To assess the value of selection and search, we compare fixed WISP methods, validation-based selection and full search. To assess the contributions of representations and temporal decoding, we conduct controlled WISP-CIS ablations. To test whether evolution improves on random search, we compare the two strategies at equal candidate budgets.

\textbf{Non-neural baselines} include Balanced-RF \citep{breiman2001rf} and RF-HMM \citep{chan2024capture}; convolutional transforms MiniRocket \citep{dempster2021minirocket}, MultiRocket \citep{tan2022multirocket} and MR-Hydra \citep{dempster2023hydra}; interval methods CIF \citep{middlehurst2020cif} and QUANT \citep{dempster2024quant}; and the shapelet method RDST \citep{guillaume2022rdst}. \textbf{Neural baselines} include supervised classifiers FCN and ResNet \citep{wang2017strong}, H-InceptionTime \citep{fawaz2022hybrid}, LITETime \citep{fawaz2025lite}, Attend-and-Discriminate \citep{abedin2021attend}, TimesNet \citep{wu2023timesnet} and PatchTST \citep{nie2023patchtst}, and self-supervised representation methods TS2Vec \citep{yue2022ts2vec}, TF-C \citep{zhang2022tfc} and HARNet \citep{yuan2024ssl}.

\label{sec:implementation}
All methods follow the splits, refitting, aggregation and metrics of Section~\ref{sec:method}, with model seeds 42, 43 and 44 per split (MiniRocket: 42 only). Appendices~\ref{app:protocol} and~\ref{app:baseline_settings} detail refitting, baseline settings and input adapters.

\subsection{Performance Against External Baselines}
\label{sec:external_results}\label{sec:main_results}

\begin{figure}[!t]
\centering
\includegraphics[width=\linewidth]{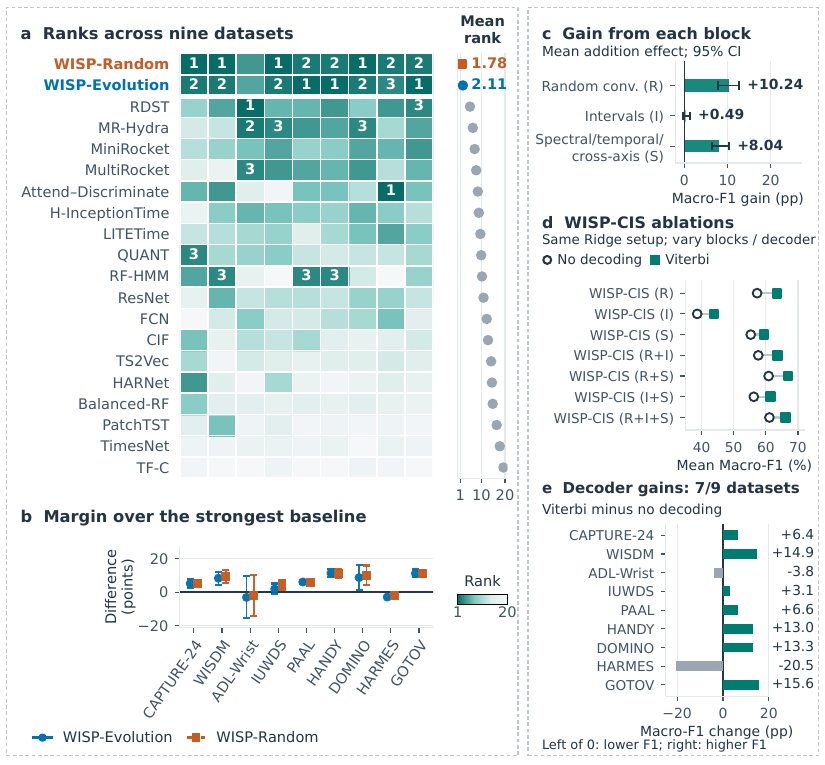}
\caption{External comparison on the nine primary datasets. (a) Macro-F1 ranks of the two WISP searches and eighteen baselines, with the three best ranks marked in each dataset and mean ranks shown at right. (b) Macro-F1 difference between each WISP search and the strongest baseline on each dataset; whiskers show the standard deviation of five paired split differences, and positive values favour WISP. (c) Mean Macro-F1 gain from adding the random convolution ($R$), interval ($I$), or spectral, temporal and cross-axis ($S$) representation, averaged over the three non-empty subsets of the other two, both decoding settings and nine datasets; error bars are 95\% dataset-bootstrap intervals. (d) Seven subsets of the WISP-CIS representations: open circles use no temporal decoding (windowwise argmax); filled squares use Viterbi decoding. Each pair differs only in decoding; $R+I+S$ is the full fixed representation. (e) Viterbi minus no-decoding Macro-F1 for each dataset, averaged over the seven subsets; pp denotes percentage points.}
\label{fig:results_composite}
\end{figure}

\begin{table}[!t]
\centering\normalsize\setlength{\tabcolsep}{3.5pt}\renewcommand{\arraystretch}{1.0}
\caption{Aggregate performance of two WISP search methods and eighteen external baselines on the nine primary datasets. Years identify the cited source papers; implementation details and input adapters are specified in Appendix~\ref{app:baseline_settings}. Scores are equal-dataset percentage means; Macro-F1 ranks and the markers bold $(1)$, underlined $(2)$ and $(3)$ are computed over all twenty methods. Per-dataset results are in Appendix~\ref{app:external_matrix}.}\label{tab:overall}
\begin{tabularx}{\linewidth}{@{}>{\raggedright\arraybackslash}Xrrrrr@{}}
\toprule
\textbf{Method} & \textbf{Mean rank} & \textbf{Macro-F1} & \textbf{Accuracy} & \textbf{Bal. acc.} & \textbf{Worst F1}\\
\midrule
\multicolumn{6}{@{}l}{\textbf{WISP search methods}}\\
\textbf{WISP-Random} (ours) & $\mathbf{1.78}^{(1)}$ & $\mathbf{70.20}^{(1)}$ & $\mathbf{77.28}^{(1)}$ & $\mathbf{70.88}^{(1)}$ & $\mathbf{28.25}^{(1)}$\\
\textbf{WISP-Evolution} (ours) & $\underline{2.11}^{(2)}$ & $\underline{69.59}^{(2)}$ & $\underline{76.84}^{(2)}$ & $\underline{70.17}^{(2)}$ & $\underline{27.96}^{(2)}$\\
\midrule
\multicolumn{6}{@{}l}{\textbf{All eighteen baselines}}\\
RDST (\citeyear{guillaume2022rdst}) & $5.11^{(3)}$ & $59.62^{(3)}$ & $69.61^{(3)}$ & $59.65$ & $17.27$\\
MR-Hydra (\citeyear{dempster2023hydra}) & $6.33$ & $59.58$ & $67.93$ & $59.59$ & $18.82$\\
MiniRocket (\citeyear{dempster2021minirocket}) & $7.11$ & $58.67$ & $68.64$ & $58.98$ & $15.35$\\
MultiRocket (\citeyear{tan2022multirocket}) & $7.78$ & $59.02$ & $67.34$ & $59.04$ & $18.99$\\
Attend-and-Discriminate (\citeyear{abedin2021attend}) & $8.44$ & $55.89$ & $62.97$ & $62.12^{(3)}$ & $22.75^{(3)}$\\
H-InceptionTime (\citeyear{fawaz2022hybrid}) & $8.89$ & $56.55$ & $65.54$ & $58.51$ & $18.04$\\
LITETime (\citeyear{fawaz2025lite}) & $9.56$ & $56.20$ & $65.42$ & $58.48$ & $18.19$\\
QUANT (\citeyear{dempster2024quant}) & $9.89$ & $55.22$ & $65.80$ & $55.16$ & $15.60$\\
RF-HMM (\citeyear{chan2024capture}) & $10.22$ & $50.39$ & $57.89$ & $55.18$ & $17.66$\\
ResNet (\citeyear{wang2017strong}) & $10.89$ & $54.42$ & $63.94$ & $56.31$ & $17.05$\\
FCN (\citeyear{wang2017strong}) & $12.22$ & $52.72$ & $61.70$ & $55.15$ & $13.90$\\
CIF (\citeyear{middlehurst2020cif}) & $12.78$ & $51.79$ & $64.62$ & $52.28$ & $11.05$\\
TS2Vec (\citeyear{yue2022ts2vec}) & $14.11$ & $50.03$ & $58.37$ & $55.16$ & $17.26$\\
HARNet (\citeyear{yuan2024ssl}) & $14.44$ & $47.90$ & $55.27$ & $52.20$ & $17.46$\\
Balanced-RF (\citeyear{breiman2001rf}) & $14.78$ & $47.38$ & $60.07$ & $48.09$ & $8.99$\\
PatchTST (\citeyear{nie2023patchtst}) & $16.44$ & $44.27$ & $50.96$ & $49.23$ & $14.22$\\
TimesNet (\citeyear{wu2023timesnet}) & $17.78$ & $43.01$ & $50.67$ & $47.24$ & $12.84$\\
TF-C (\citeyear{zhang2022tfc}) & $19.33$ & $34.66$ & $41.35$ & $38.76$ & $8.76$\\
\midrule
\multicolumn{6}{@{}l}{\textbf{Fixed references (excluded from ranks above)}}\\
WISP-CIS, direct (ours) & --- & $61.12$ & $68.81$ & $66.42$ & $22.17$\\
WISP-CIS, $H$ (ours) & --- & $66.14$ & $74.07$ & $66.04$ & $24.78$\\
\bottomrule\end{tabularx}

\end{table}

Across the nine primary datasets, WISP-Random and WISP-Evolution rank first and second across all four aggregate metrics (Table~\ref{tab:overall}; Figure~\ref{fig:results_composite}a). Their mean Macro-F1 scores are 70.20\% and 69.59\%, compared with 59.62\% for RDST, the strongest external baseline on this metric. Both searches exceed the strongest baseline on seven datasets, with the comparator selected separately per dataset; ADL-Wrist and HARMES favour RDST and Attend-and-Discriminate, respectively (Figure~\ref{fig:results_composite}b). The strongest baseline itself changes across datasets: convolutional and shapelet transforms lead on ADL-Wrist, DOMINO, GOTOV and IUWDS, whereas RF-HMM leads on HANDY, PAAL-ADL and WISDM-watch. These datasets also differ in sampling rate (20--100\,Hz), number of activity classes (4--24) and number of participants (16--151) (Table~\ref{tab:datasets}), so no single representation and classifier is favoured throughout. Fixed WISP-CIS with temporal decoding also exceeds the strongest external aggregate score on all four metrics without recipe search.

These results are consistent with the representation hypothesis in Section~\ref{sec:intro}: wrist activities differ in complementary motion patterns, including local shape, rhythm, temporal order and cross-axis relationships (Figure~\ref{fig:wisp_overview}). Among the external baselines, MiniRocket, MultiRocket, MR-Hydra and RDST all represent local shape, using random kernels, multiple pooling operators, convolutional pattern counts and dilated shapelets, respectively \citep{dempster2021minirocket,tan2022multirocket,dempster2023hydra,guillaume2022rdst}. Their mean Macro-F1 scores lie within one percentage point of one another (58.67--59.62\%), and the random convolution representation of WISP-CIS alone reaches 57.28\% with direct prediction. Different representations of local shape therefore reach similar scores. Adding the spectral, temporal and cross-axis representation, which describes rhythm, temporal order and cross-axis relationships, raises performance above this range. With the classifier fixed, $R+S$ exceeds the better of $R$ and $S$ on each of the nine primary datasets with direct prediction and on eight with temporal decoding. Adding $R$ to $I+S$, or $S$ to $R+I$, still increases Macro-F1 by 4.80 and 3.01 percentage points, respectively (Figure~\ref{fig:results_composite}c--d). Because these gains occur within individual datasets, the two representations capture complementary patterns rather than merely suiting different datasets. When one representation fits a dataset particularly well, fixed WISP-CIS can trail it. WISP-SS reaches 96.42\% on HANDY and 75.62\% on PAAL-ADL, compared with 89.63\% and 71.98\% for WISP-CIS, whereas WISP-Random reaches 95.77\% and 76.45\%.

\subsection{Representation, Decoding and Search}
\label{sec:internal_results}

\paragraph{Representation component ablations.}
\label{sec:ingredient_results}
Starting from fixed WISP-CIS ($RIS+H$), we evaluate all seven non-empty combinations of $R$, $I$ and $S$, each with direct prediction or Viterbi decoding, giving fourteen configurations (Figure~\ref{fig:results_composite}d). The classifier type, settings of retained representations, scaling procedure and split/seed protocol remain fixed. Each representation's average contribution is the mean Macro-F1 change from adding it to the three non-empty combinations of the other two under both prediction modes. We also evaluate its contribution when both others are present \mbox{(Appendix~\ref{app:internal_results})}. Adding the random convolution representation ($R$) and the spectral, temporal and cross-axis representation ($S$) gives mean Macro-F1 gains of 10.24 and 8.04 percentage points, respectively (Figure~\ref{fig:results_composite}c). For both representations, the average gain is positive on all nine datasets. The gain from each representation depends on which patterns are already present. Adding $R$ to $I$ alone increases Macro-F1 by 19.0--19.7 percentage points across the two prediction modes, but by 4.7--4.9 points once $S$ is present. The gain from $S$ falls similarly, from 17.5--17.6 points over $I$ alone to 2.6--3.5 points over $R+I$ (Figure~\ref{fig:results_composite}d). $R$ and $S$ therefore partly overlap in the patterns they capture, while each still adds patterns that the other lacks. The components within $S$ are not evaluated separately.

Adding the interval representation ($I$) gives an average Macro-F1 gain of 0.49 percentage points. With both $R$ and $S$ present, the gain averaged over both prediction modes is $-0.27$ points; both 95\% confidence intervals include zero. The average accuracy gain is 0.58 percentage points (Appendix~\ref{app:internal_results}). Once $R$ is present, adding $I$ changes Macro-F1 by $-0.79$ to $+0.37$ points under either prediction mode; over $S$ alone, it adds at most 2.03 points. A plausible explanation is that $I$ adds few new patterns. It summarises value distributions within intervals, whereas $R$ already pools the proportion of positive values, mean, standard deviation and maximum of each kernel response, and $S$ records occupancies of quantile states over the window (Section~\ref{sec:representation}). Among the fourteen configurations, $RS+H$ achieves the highest observed mean Macro-F1 at 66.93\%, 0.79 percentage points above full WISP-CIS. This is a post-hoc test comparison; the best configuration varies by dataset. These results also suggest that a representation improves recognition when it adds patterns that the others lack, not merely because the number of representations increases.

\paragraph{Temporal decoding ablation.}
\label{sec:decoder_results}
Temporal decoding improves mean Macro-F1 on seven datasets and reduces it on ADL-Wrist and HARMES (Figure~\ref{fig:results_composite}e). Averaged over seven representations and nine datasets, the gain is 5.39 percentage points (95\% CI: $-2.39$ to $11.65$). On HARMES, adding decoding to fixed $RIS$ changes Macro-F1 from 44.43\% to 23.99\%; Random reaches 44.84\%. Decoding assumes that neighbouring windows usually belong to the same activity (Appendix~\ref{app:heads}). This assumption corrects isolated window errors, but when window predictions are weak it can merge different activities. On HARMES, decoding leaves only 7.3 of 16 activities with at least 1\% of predictions, averaged over the seven representations; on ADL-Wrist, 9.7 of 14 remain, compared with 13.3 without decoding. These two datasets also have the lowest direct-prediction Macro-F1. Decoding therefore helps when activity sequences match its assumption.

\label{sec:family_results}WISP-Select5 and WISP-Select7 achieve 67.10\% and 66.69\% mean Macro-F1, respectively, compared with 66.14\% for fixed WISP-CIS. Choosing the best of the five fixed representations on test data would give 67.69\%, so validation-based selection recovers 0.96 of the possible 1.55 points. Selection helps most when one representation fits a dataset well: on HANDY, Select5 reaches 96.31\%, close to WISP-SS (96.42\%) and 6.68 points above WISP-CIS. The full searches (70.20\% and 69.59\%) also change the decoder and within-family settings. Their additional gain is concentrated on HARMES. There, Random exceeds Select5 by 21.33 points; on the other eight datasets, the average margin is 0.82 points. Both selectors keep decoding enabled, whereas Random disables it in all 30 selected HARMES configurations (Appendix~\ref{app:search_concentration}). Adapting to a dataset therefore involves choosing both the representation and whether to decode.
\ifdefined\ExtendedMode\ifdefined\ExtendedMode\begin{table}[t]\else\begin{table}[H]\fi
\centering\ifdefined\ExtendedMode\normalsize\else\small\fi\setlength{\tabcolsep}{3.5pt}\renewcommand{\arraystretch}{1.02}
\caption{Internal comparison of eleven WISP methods. Scores are equal-dataset percentage means over the primary nine. Ranks use these eleven entries; the two complete methods share their measurements with Table~\ref{tab:overall}. First, second and third are marked as in that table. Per-dataset Macro-F1 results are in Appendix~\ref{app:internal_results}.}\label{tab:wisp_internal}
\begin{tabularx}{\linewidth}{@{}Xrrrrr@{}}
\toprule
\textbf{Method} & \textbf{Mean rank} & \textbf{Macro-F1} & \textbf{Accuracy} & \textbf{Bal. acc.} & \textbf{Worst F1}\\
\midrule
\multicolumn{6}{@{}l}{\textbf{Full methods: shared reference points}}\\
\textbf{WISP-Random} (ours) & $\mathbf{2.78}^{(1)}$ & $\mathbf{70.20}^{(1)}$ & $\mathbf{77.28}^{(1)}$ & $\mathbf{70.88}^{(1)}$ & $\mathbf{28.25}^{(1)}$\\
\textbf{WISP-Evolution} (ours) & $\underline{3.11}^{(2)}$ & $\underline{69.59}^{(2)}$ & $\underline{76.84}^{(2)}$ & $\underline{70.17}^{(2)}$ & $\underline{27.96}^{(2)}$\\
\addlinespace[3pt]
\multicolumn{6}{@{}l}{\textbf{Fixed representation and fusion families}}\\
WISP-SS (ours) & $6.89$ & $58.54$ & $73.36$ & $59.04$ & $21.70$\\
WISP-RC (ours) & $7.00$ & $63.44$ & $71.46$ & $63.59$ & $22.87$\\
WISP-SO (ours) & $10.33$ & $53.59$ & $69.31$ & $54.19$ & $16.75$\\
WISP-CSE (ours) & $6.44$ & $58.74$ & $72.86$ & $59.10$ & $21.22$\\
WISP-GIS (ours) & $9.33$ & $55.16$ & $70.66$ & $55.95$ & $19.49$\\
WISP-CIS (ours) & $4.44$ & $66.14$ & $74.07$ & $66.04$ & $24.78$\\
WISP-ESE (ours) & $8.00$ & $57.38$ & $71.73$ & $57.71$ & $19.78$\\
\addlinespace[3pt]
\multicolumn{6}{@{}l}{\textbf{Validation selection}}\\
WISP-Select5 (ours) & $3.44^{(3)}$ & $67.10^{(3)}$ & $75.21^{(3)}$ & $66.94^{(3)}$ & $24.93^{(3)}$\\
WISP-Select7 (ours) & $4.22$ & $66.69$ & $75.01$ & $66.65$ & $24.65$\\
\bottomrule\end{tabularx}

\end{table}
\fi

\paragraph{Search strategy and budget.}
\label{sec:budget_results}
At candidate budgets of 25, 50, 100, 200 and 329, we select the validation-best recipe from each recorded search prefix before test evaluation. No budget shows an evolutionary advantage after Holm adjustment over the five comparisons; at 200 and 329 candidates, Random is 0.53 and 0.61 points higher, with unadjusted 95\% intervals excluding zero (Appendix Table~\ref{tab:budget_tests}). The normalised log-budget area difference (Evolution minus Random) is $-0.09$ Macro-F1 points, with 95\% interval [$-0.52$, +0.31]; the other metrics likewise show no reliable evolutionary advantage. After 25 candidates, both searches are already within 1.1 points of their final mean Macro-F1 (Appendix~\ref{app:budget_curves}). This suggests that strong recipes are common in the WISP space, leaving little room for earlier validation scores to guide the search, consistent with earlier evidence on random search \citep{bergstra2012random,li2019random}. Performance thus depends more on the representations the space contains than on how it is searched.

\section{Conclusion}
\label{sec:practical}
We introduced \bench{}, a participant-disjoint benchmark for wrist activity recognition, and WISP, a composable representation space for wrist motion patterns. \textbf{Different representations of local shape reach similar performance, whereas adding representations of rhythm, temporal order and cross-axis relationships improves recognition on every primary dataset}; representations that repeat patterns already present add little. \textbf{Wrist activity recognition may therefore benefit more from representing patterns that current representations omit than from new representations of the same pattern.} The most useful representation still varies across datasets, and temporal decoding helps on most datasets but can merge activities on others, so both are best chosen for each dataset.

\FloatBarrier\label{maintextend}

\section*{Acknowledgments}
F.S. was supported by a UKRI Future Leaders Fellowship, grant numbers [MR/T043571/1] and [MR/Z000262/1]. The computations described in this paper were performed using the University of Birmingham's BlueBEAR HPC service, which provides a High Performance Computing service to the University's research community. See \url{http://www.birmingham.ac.uk/bear} for more details.

\section*{AI use statement}
ChatGPT and Claude were used to improve the wording, organisation and LaTeX presentation of author-written text, to help write and check the code for the methods, experiments and analyses, and to check reported numbers against saved result files. All reported measurements come from executed experiments. The authors take responsibility for all AI-assisted text and code and for the final content of the manuscript.

\section*{Reproducibility statement}
Section~\ref{sec:method} defines the representation space and evaluation protocol. Appendix~\ref{app:implementation} summarises data export, representation operations, configurations, search and scoring, and the released code gives the complete implementation; Appendices~\ref{app:complete_results} and~\ref{app:internal_results} give complete results and supporting analyses. Code and saved results are available at \url{https://huggingface.co/Zipeng365/WISP}, and the benchmark data at \url{https://huggingface.co/datasets/Zipeng365/WristHARBench}.

\section*{Ethics statement}
This study analyses existing wearable datasets and introduces no new human-participant collection. Dataset access conditions and licences remain those of their providers. Activity-recognition performance does not establish clinical benefit, and population-specific errors require evaluation before use in health decisions.

\bibliographystyle{preprint}
\bibliography{references}
\appendix
\section{Additional execution details}
\label{app:implementation}\label{app:extended_design}

\subsection{Fit, select, refit and evaluate}
\label{app:protocol}

Split seeds are 1000/2000/3000/4000/5000. Representations, scaling, classifiers and $H$ are fitted on training data. Select5/7 use validation Macro-F1; Random/Evolution use $U$ (Appendix~\ref{app:search}). Fixed recipes and selected candidates are refitted from scratch on training plus validation data before test evaluation.

\subsection{Dataset export rules}
\label{app:export_rules}

For CAPTURE-24, at most 500 approximately evenly spaced candidate windows per participant are drawn before label filtering. Table~\ref{tab:export_rules} lists candidate strides; filtering, gaps and activity grouping determine the spacing of retained windows.

\begin{longtable}{L{21mm}L{23mm}L{81mm}}
\caption{Dataset export rules. Window lengths and strides are in seconds. For CAPTURE-24 and IUWDS, the label fraction is the proportion of valid point annotations supporting the assigned class.}\label{tab:export_rules}\\
\toprule Dataset & Window/stride & Label assignment and source mapping\\\midrule\endfirsthead
\multicolumn{3}{l}{\textit{Table~\thetable{} continued}}\\
\toprule Dataset & Window/stride & Label assignment and source mapping\\\midrule\endhead\bottomrule\endfoot
CAPTURE-24 & 10/10 & Walmsley2020 labels of the CAPTURE-24 release \citep{chan2024capture}; majority valid annotation with fraction at least .8.\\
WISDM & 5/2.5 & Official activity-code mapping; windows constructed within each activity-code group, ordered by timestamp.\\
ADL-Wrist & 5/2.5 & Activity directory supplies the trial label; model-template directories excluded.\\
IUWDS & 5/2.5 & Majority point label, fraction at least .8; background code 99 retained; acquisition gaps above $0.05$~s split source segments.\\
PAAL-ADL & 5/2.5 & Activity, participant and repetition parsed from trial names; trials shorter than 5~s receive end-edge padding.\\
HANDY & 5/2.5 & Participant/activity filenames mapped to ten classes.\\
DOMINO & 5/2.5 & Named activity intervals; TRANSITION excluded; linear interpolation to 100~Hz within sensor chunks, splitting gaps above 1~s.\\
HARMES & 5/5 & Maximum-overlap interval label including Null; protocol recordings 01--03.\\
GOTOV & 5/2.5 & Sixteen mapped activities, unknown labels excluded; interpolation to 83~Hz within constant-label segments, splitting gaps above 100~ms; stride rounded to 208 samples.\\
PAMAP2 & 5/2.5 & Official activity-ID mapping; ID 0 excluded; windows constructed within each activity-ID group.\\
MHEALTH & 5/2.5 & Official activity-ID mapping; ID 0 excluded; windows constructed within each activity-ID group.\\
\end{longtable}

\subsection{Deterministic and sampled representation operations}
\label{app:operators}\label{app:sampled}\label{app:random}

Equation~\ref{eq:window_standardisation} uses NaN-omitting means and population standard deviations; non-finite outputs become zero. Set $\Delta\widetilde X_{1,:}=0$ and $\Delta\widetilde X_{t,:}=\widetilde X_{t,:}-\widetilde X_{t-1,:}$ for $t>1$, without time normalisation. Magnitudes $m_t,j_t$ are the Euclidean norms of the first $\min(C,3)$ channels of $\widetilde X$ and $\Delta\widetilde X$. Global statistics per scalar view are mean, population standard deviation, extrema, median, quartiles, mean square, first-difference mean/standard deviation, zero-crossing rate, least-squares slope, skewness and kurtosis. Equal-length segments use end-value padding and contribute mean, population standard deviation, range and least-squares slope.

For mean-centred $v$, spectral power is $|\operatorname{rFFT}(v)|^2/T$. Bands $[0,.05)$/$[.05,.15)$/$[.15,.30)$/$[.30,.51)$ cycles/sample are normalised by non-DC power plus $10^{-12}$. Non-DC entropy is divided by the log bin count (minimum two); the power-weighted centroid and dominant frequency also exclude DC. Autocorrelation uses lags from $\{1,2,4,8,16,32\}$ below $T$, divided by $\sum_t(v_t-\bar v)^2+10^{-12}$. Correlations are pairwise Pearson correlations of standardised channels. State statistics (Equation~\ref{eq:state_order}) use at most the first three channels per enabled view: occupancies, flattened $Q$ and $-\sum_{ab}Q_{ab}\log(Q_{ab}+10^{-12})$.

For random convolution, uniformly draw $L\in\{7,9,11,15\}$, $d\in\{1,2,4\}$ and a channel count in $1,\ldots,\min(C,3)$; sample channels without replacement and sort them. If $(L-1)d+1>T$, set $d=\max(1,\lfloor(T-1)/(L-1)\rfloor)$. Centre independent standard Gaussian weights per channel, divide by their joint Euclidean norm plus $10^{-12}$, and draw bias from $\mathcal N(0,.25^2)$. Reuse the fitted kernel bank at validation and test.

Consecutive fixed intervals use widths $\max(2,\lceil T/p\rceil)$ for $p\in\{2,4,8\}$, starting at zero; clip at $T$ and discard lengths below two. Random intervals draw start $a$ uniformly from $0,\ldots,\max(T-2,1)-1$ and length uniformly from $2,\ldots,\max(3,T-a)$, clipping at $T$ and deduplicating in order. Each interval/channel returns mean, population standard deviation, range, mean square, least-squares slope and .10/.25/.50/.75/.90 quantiles.

For GIS subsequences, uniformly draw $L\in\{8,12,16,24\}$, $d\in\{1,2,4\}$ (clipped as above), then training-window index, channel and valid start. Centre prototypes and divide by population standard deviation (zero replaced by one); matching segments in the same channel use the same normalisation with deviations below $10^{-6}$ replaced by one. Return minimum RMS distance and the first attaining zero-based start divided by $\max(T-(L-1)d-1,1)$.

GIS uses $C_3=\min(C,3)$, $V=\widetilde X_{:,1:C_3}$ and the temporal mean $\bar V$ of its rows to compute
\begin{equation}
 G(X)=\frac{1}{\max(T-1,1)}\sum_{t=1}^{T}(V_{t,:}-\bar V)^{\mathsf T}(V_{t,:}-\bar V).
\label{eq:geometry_matrix}
\end{equation}
It clips negative eigenvalues of $G(X)$ to zero, sorts descending, appends ratios to their sum plus $10^{-12}$, and zero-pads to six coordinates; it adds $\mathbb E[j^2]/(\mathbb E[m^2]+10^{-12})$. Its axis branch uses global and segment statistics and correlations; the magnitude and difference-magnitude views also use spectral, autocorrelation and state statistics. The $S$ representation of CIS uses spectral, autocorrelation, six-state and cross-axis correlation statistics, without global or segment statistics.

\subsection{Fixed recipes, prediction heads and admissible changes}
\label{app:heads}\label{app:family_settings}

A WISP recipe $r=(g,\theta,h,d,w,\xi)$ specifies the method $g$, representation settings $\theta$, classifier settings $h$, decision rule $d\in\{\mathrm{direct},H\}$, ensemble weights $w$ where applicable, and random seed $\xi$; the allowed settings depend on the method.

ExtraTrees/random forest \citep{pedregosa2011scikit} use 120 trees (SO: 160 ExtraTrees), square-root feature sampling, minimum leaf size one, and balanced/balanced-subsample weights, respectively. Logistic regression uses $C_{\mathrm{LR}}=2$, L-BFGS, at most 2,000 iterations and balanced weights. Logistic/Ridge inputs are standardised on the outer training set. \texttt{RidgeClassifierCV} uses balanced sample weights and nine log-spaced penalties $10^{-3}$--$10^3$; default window-level leave-one-out/GCV minimises squared error after outer-set scaling, without participant-grouped inner folds.

Class scores use native probabilities, else softmax margins (binary: $[-d,d]$), else one-hot predictions. Align estimator class IDs to the dataset vocabulary, zero absent fitted classes and normalise rows (all-zero rows become uniform). Ensemble members use seeds $\xi,\xi+1,\xi+2$ with optional $H$ applied only after fusion.

For $H$, count adjacent training labels within participants, add one to all transition counts and two more on the diagonal, then row-normalise. Initial class probabilities use training counts plus one. Floor scores at $10^{-12}$, renormalise and run log-space Viterbi on each participant's full sequence, without recording resets. Windows are ordered by start time, or by row order when no start time is recorded. Fitting/refitting follows Appendix~\ref{app:protocol}.

\begin{table}[htbp]\centering\small\setlength{\tabcolsep}{3pt}
\caption{Fixed recipes. Every recipe uses final Viterbi decoding $H$; member decoding is disabled in ensembles.}\label{tab:family_settings}
\begin{tabularx}{\linewidth}{@{}L{.15\linewidth}X@{}}
\toprule Family & Configuration\\\midrule
WISP-SS & Four views; all shared statistics; segments 4/8/16; six states; ExtraTrees.\\
WISP-RC & 384 kernels; Ridge.\\
WISP-SO & Axes and magnitude; global/5/10/20 segments; correlations; eight states; 160 trees.\\
WISP-CSE & SS, RC (256 kernels), SO; equal weights.\\
WISP-GIS & Axis and geometry branches; 24 random intervals; 32 subsequences; ExtraTrees.\\
WISP-CIS & 384 kernels; 24 random intervals; six-state $S$ representation; Ridge.\\
WISP-ESE & CIS (256 kernels, 16 intervals), GIS (24 subsequences), SO; weights .34/.33/.33 (search initialisation: .40/.35/.25).\\
\bottomrule\end{tabularx}
\end{table}

All families expose $H$ and seed to search. SS/GIS also expose head, segments and states; GIS adds subsequence count. RC/CIS expose kernel count and Ridge/logistic heads; CIS adds interval count. ESE also exposes fusion weights. Controlled CIS comparisons toggle $R/I/S$.

Search ranges: three sorted segment counts sampled without replacement from $\{3,\allowbreak 4,\allowbreak 5,\allowbreak 8,\allowbreak 10,\allowbreak 12,\allowbreak 16,\allowbreak 20,\allowbreak 25\}$; states $\{4,\allowbreak 5,\allowbreak 6,\allowbreak 8,\allowbreak 10\}$; kernels $\{128,\allowbreak 256,\allowbreak 384,\allowbreak 512,\allowbreak 768\}$; intervals $\{8,\allowbreak 16,\allowbreak 24,\allowbreak 32,\allowbreak 48\}$; subsequences $\{16,\allowbreak 24,\allowbreak 32,\allowbreak 48,\allowbreak 64\}$. A small-sample descriptor weight of at least .6 (Appendix~\ref{app:search}) restricts kernels to $\{128,\allowbreak 256,\allowbreak 384\}$ and subsequences to $\{8,\allowbreak 16,\allowbreak 24\}$. Ensemble weights follow $\operatorname{Dirichlet}(1,1,1)$.

\subsection{Search specification}
\label{app:search}

For validation Macro-F1 $m$, worst-class F1 $w_{\min}$, fit/prediction times $t_f,t_p$ and mean Macro-F1 $m_s$ on the four perturbed validation inputs defined below, search uses
\begin{align}
 c(r)&=n_{\mathrm{op}}+n_k/256+n_s/64,\nonumber\\
 q(r)&=0.75m+0.25w_{\min}-0.01\log(1+t_f+2t_p)-0.002c(r),\label{eq:q}\\
 U(r)&=q(r)+0.10m_s-0.05\max(0,m-m_s).\label{eq:utility}
\end{align}
$n_{\mathrm{op}}$ counts stored operator labels; $n_k,n_s$ are top-level kernel/subsequence settings (zero if absent), not ensemble totals. Validation $m,w_{\min}$ use the true/predicted-label union; test worst-class F1 uses the full vocabulary. Scores are fractions, times seconds; coefficients are fixed heuristics. The fitted model is reused under all perturbations.

Nine initial recipes (seed offsets 0--8) are SS/ExtraTrees (4/8/16 segments, six states), SS/random forest (5/10/20, eight), RC/Ridge (256 kernels), RC/logistic (512), SO, CSE, GIS, CIS and ESE. All use $H$ except RC/logistic; other settings follow Table~\ref{tab:family_settings}. Initial preferences over families and changes come from hand-defined rules applied to training-data descriptors: sample, participant and device counts, dimensions, imbalance, spectral concentration, derivative burstiness, lag-one dependence, channel correlation, missingness, sampling rate and study setting. Descriptors use at most 4,096 training windows and are recomputed for each run, without cross-dataset transfer.

Random applies one to four admissible changes to a freshly seeded template, without past scores. Evolution groups parents by family, cost category and operator labels; with probability .25 it samples a smallest group, otherwise a uniformly chosen group, and within the group it samples a parent's breeding-score rank geometrically ($p=.55$, capped at the last rank). Each round evaluates sixteen candidates and retains the top 24 accepted parents and offspring.

Mutation preferences add $10\max(0,\overline U_m-\operatorname{median}_{m'}\overline U_{m'})$, using each mutation's mean recorded utility; clip weights below zero, add $10^{-6}$ and normalise over admissible changes. Breeding scores add $.02$ times the mean Jaccard dissimilarity to the latest fifty operator-label sets; final selection uses $U$. Duplicates are identified by normalised family, settings, operator labels and parent lineage, so different seeds can specify the same structure. Perturbations are: reverse axes one and three ($C\geq3$); circularly shift by $\max(1,\lfloor T/20\rfloor)$ samples; zero the final channel ($C>1$); add Gaussian noise with standard deviation 5\% of each window's standard deviation across samples/channels (seed 12345). Failed/rejected candidates, including non-finite scores or collapsed multiclass predictions, consume budget.

Budgets $25,50,100,200,329$ are prefixes of the same saved searches. Select maximum validation $U$, breaking ties by shorter fit time, before accessing test arrays; identical winners are refitted once across budgets. Over the 150 primary search runs per method (nine datasets with HARMES wrists counted separately, five splits and three seeds), the median cumulative candidate fitting time is 61.3~h for WISP-Evolution and 47.5~h for WISP-Random on CPUs; neural baselines ran on GPUs, so timings are not hardware-matched.

\subsection{Scoring and uncertainty}
\label{app:metric_definitions}\label{app:split_sizes}

For confusion matrix $M$ on the dataset vocabulary, set $n_k=\sum_jM_{kj}$, $\widehat n_k=\sum_jM_{jk}$ and $F_k=2M_{kk}/(n_k+\widehat n_k)$ (zero if the denominator is zero). Accuracy is $\operatorname{tr}(M)/\sum_{kj}M_{kj}$; Macro-F1 averages $F_k$ over $n_k+\widehat n_k>0$; balanced accuracy averages $M_{kk}/n_k$ over $n_k>0$; worst-class F1 takes the full-vocabulary minimum, including absent classes. Average seeds within each stream, pair HARMES wrists within split, then report five-split mean and sample standard deviation ($\mathrm{ddof}=1$). Bootstrap intervals and exact sign-flip tests use nine paired dataset differences, with Holm adjustment within each test group; repeated splits are not independent cohorts.

Several descriptive quantities in Section~\ref{sec:experiments} are computed after all runs from saved test outputs and are not used for selection. Gains of a representation in a given context are differences between the nine-dataset means of the corresponding rows of Table~\ref{tab:rev6_internal_macro_f1}. The number of activities receiving at least 1\% of predictions uses confusion matrices summed over splits, seeds and both HARMES wrists, averaged over the seven representation subsets. The test-selected reference (67.69\%) takes, for each dataset, the highest test Macro-F1 among the five fixed recipes available to WISP-Select5.

\newcommand{\wispinternalheading}{\section{Complete internal representation and search analysis}\label{app:internal_results}\label{app:internal_series_matrix}}
\newcommand{\wispinternalmatrixscope}{The Macro-F1 matrix covers all eleven datasets, separately ranking eleven complete designs and fourteen CIS recipes; $RIS+H$ is fixed~CIS. Per-dataset internal results for the other three metrics are included with the released results.}
\section{Complete comparison with existing methods}\label{app:complete_results}\label{app:external_matrix}

\subsection{Baseline settings and paired comparisons}
\label{app:baseline_settings}

Table~\ref{tab:baseline_settings} specifies the evaluated configurations and input adapters. In Table~\ref{tab:baseline_tests}, each search exceeds fourteen baselines on all nine primary datasets and MultiRocket, MR-Hydra, RDST and Attend-and-Discriminate on eight. With nine paired differences, the smallest attainable two-sided exact $p$-value is $2/512\approx0.0039$; Holm adjustment over eighteen tests gives $p_H=0.070$ for every contrast, the smallest attainable adjusted value. Bootstrap intervals are unadjusted.

\begin{table}[H]\centering\small\setlength{\tabcolsep}{3pt}
\caption{Paired Macro-F1 contrasts against all eighteen baselines over nine dataset means. Differences and 95\% dataset-bootstrap intervals are percentage points; $p_H$ is the exact sign-flip $p$-value with Holm adjustment within each set of eighteen tests.}\label{tab:baseline_tests}
\begin{tabularx}{\linewidth}{@{}Xrrr rrr@{}}
\toprule & \multicolumn{3}{c}{WISP-Random minus baseline} & \multicolumn{3}{c}{WISP-Evolution minus baseline}\\
Baseline & Difference & 95\% interval & $p_H$ & Difference & 95\% interval & $p_H$\\\midrule
RDST & +10.58 & [+5.62, +15.33] & 0.070 & +9.97 & [+4.77, +15.01] & 0.070\\
MR-Hydra & +10.62 & [+6.36, +14.55] & 0.070 & +10.01 & [+5.47, +14.26] & 0.070\\
MiniRocket & +11.53 & [+6.88, +16.36] & 0.070 & +10.93 & [+5.99, +16.07] & 0.070\\
MultiRocket & +11.18 & [+6.78, +15.20] & 0.070 & +10.57 & [+5.94, +14.87] & 0.070\\
Attend-and-Discriminate & +14.31 & [+8.51, +20.16] & 0.070 & +13.70 & [+7.89, +19.44] & 0.070\\
H-InceptionTime & +13.65 & [+9.01, +17.96] & 0.070 & +13.04 & [+8.09, +17.71] & 0.070\\
LITETime & +14.00 & [+9.31, +18.57] & 0.070 & +13.39 & [+8.47, +18.27] & 0.070\\
QUANT & +14.98 & [+10.01, +20.00] & 0.070 & +14.37 & [+9.16, +19.74] & 0.070\\
RF-HMM & +19.81 & [+11.18, +31.18] & 0.070 & +19.20 & [+10.90, +30.22] & 0.070\\
ResNet & +15.78 & [+11.75, +19.71] & 0.070 & +15.17 & [+10.82, +19.49] & 0.070\\
FCN & +17.48 & [+12.31, +21.95] & 0.070 & +16.87 & [+11.49, +21.67] & 0.070\\
CIF & +18.41 & [+12.82, +23.83] & 0.070 & +17.81 & [+12.11, +23.38] & 0.070\\
TS2Vec & +20.17 & [+15.26, +25.09] & 0.070 & +19.57 & [+14.53, +24.71] & 0.070\\
HARNet & +22.30 & [+15.65, +28.65] & 0.070 & +21.69 & [+14.86, +28.28] & 0.070\\
Balanced-RF & +22.82 & [+17.35, +28.11] & 0.070 & +22.21 & [+16.59, +27.80] & 0.070\\
PatchTST & +25.93 & [+19.78, +32.08] & 0.070 & +25.32 & [+18.99, +31.72] & 0.070\\
TimesNet & +27.19 & [+21.61, +32.93] & 0.070 & +26.58 & [+20.82, +32.51] & 0.070\\
TF-C & +35.54 & [+28.91, +42.55] & 0.070 & +34.93 & [+28.01, +42.28] & 0.070\\
\bottomrule\end{tabularx}\end{table}

{\small
\begin{longtable}{L{27mm}L{103mm}}
\caption{Baseline configurations and input adapters.}\label{tab:baseline_settings}\\
\toprule Method & Configuration and input adapter\\\midrule\endfirsthead
\multicolumn{2}{l}{\textit{Table~\thetable{} continued}}\\
\toprule Method & Configuration and input adapter\\\midrule\endhead\bottomrule\endfoot
Balanced-RF & \citep{breiman2001rf}. 300 trees, balanced class weights, model seed varied per run.\\
RF-HMM & \citep{chan2024capture}. CAPTURE-24 author-source extractor and HMM; 3,000 balanced-forest trees, replacement sampling, sampling strategy ``not minority'', square-root split sampling. First three acceleration axes; the extractor runs at each dataset's sampling rate. Inputs are axis-standardised, so amplitude features are not in physical units.\\
MiniRocket & \citep{dempster2021minirocket}. Author multivariate transform; 10,000 outputs; maximum 32 dilations/kernel; Ridge head; transformation batches 2,048; seed 42 only.\\
MultiRocket, MR-Hydra & \citep{tan2022multirocket,dempster2023hydra}. Pinned aeon 1.5.0 package defaults; seeds 42/43/44. MR-Hydra uses the package's multivariate extension.\\
CIF, QUANT & \citep{middlehurst2020cif,dempster2024quant}. aeon 1.5.0 package defaults. CIF adds a zero-mean alternating offset of $\pm10^{-5}$ to near-constant channels and QUANT scales inputs by 16, so that aeon's low-variation input checks pass; estimator settings are unchanged.\\
RDST & \citep{guillaume2022rdst}. aeon 1.5.0 package defaults; 10,000 shapelets.\\
FCN, ResNet & \citep{wang2017strong}. aeon 1.5.0; maximum 50 epochs; batch size 64; patience five on training loss; model seeds 42/43/44.\\
H-InceptionTime & \citep{fawaz2022hybrid}. Same 50-epoch/batch-64 schedule; custom filters enabled; one classifier; depth six; 32 filters.\\
LITETime & \citep{fawaz2025lite}. Same 50-epoch/batch-64 schedule; multivariate option enabled; one classifier; 32 filters.\\
Attend-and-Discriminate & \citep{abedin2021attend}. Author-source model; hidden size 128, 64 filters of size five, two unidirectional recurrent layers; 300 epochs, batch 256, learning rate $10^{-3}$; decay .9 every ten epochs; mixup .8; centre-loss weight $3\times10^{-4}$.\\
TimesNet & \citep{wu2023timesnet}. Classification backend; model dimension 128, feedforward dimension 256, two layers, top-$k=3$, six kernels, dropout .1; 24 epochs, batch 512, learning rate $5\times10^{-4}$, weight decay $10^{-4}$, patience five.\\
PatchTST & \citep{nie2023patchtst}. Classification backend; model dimension 128, feedforward dimension 256, two layers, four heads, patch length 16, stride 8, dropout .1; same 24-epoch/batch-512 schedule.\\
TS2Vec & \citep{yue2022ts2vec}. Dataset-trained encoder; output dimension 128, hidden dimension 64, depth four; batch 16, learning rate $10^{-3}$; full-series encoding and downstream classifier.\\
TF-C & \citep{zhang2022tfc}. 40 pretraining and 40 fine-tuning epochs, batch 128, learning rate $3\times10^{-4}$, temperature .2. The input adapter supplies only the first standardised channel.\\
HARNet & \citep{yuan2024ssl}. Pretrained \texttt{harnet10}; first three acceleration axes; encoder parameters frozen, head trained up to 20 epochs, batch 64, learning rate $10^{-3}$, weight decay $10^{-4}$, patience five. Inputs are resampled to 30 Hz, then truncated or zero-padded to 300 points; a five-second window occupies about 150 points. Batch-normalisation running statistics are updated during head training.\\
\end{longtable}}

\begin{landscape}
The four matrices report all eleven datasets, extending the primary-nine aggregate comparison in Table~\ref{tab:overall}.
\begin{table}[H]
\centering\fontsize{8.8}{10.0}\selectfont\setlength{\tabcolsep}{2pt}\renewcommand{\arraystretch}{1.0}
\caption{External comparison of two WISP searches and eighteen baselines: Macro-F1 percentages (mean $\pm$ standard deviation across five split means). Columns are datasets; starred tasks are additional analyses. Bold (1), underline (2) and (3) mark the top three unrounded scores within each dataset and comparison group; ties share rank.}\label{tab:rev6_external_macro_f1}
\begin{tabularx}{\linewidth}{@{}Xrrrrrrrrrrr@{}}
\toprule Method / recipe & CAPTURE-24 & WISDM & ADL & IUWDS & PAAL & HANDY & DOMINO & HARMES & GOTOV & PAMAP2* & MHEALTH*\\\midrule
\textbf{WISP-Random} & $\mathbf{74.4}^{(1)}\,{\scriptscriptstyle\pm}\,1.7$ & $\mathbf{74.0}^{(1)}\,{\scriptscriptstyle\pm}\,6.3$ & $54.2\,{\scriptscriptstyle\pm}\,13.7$ & $\mathbf{63.2}^{(1)}\,{\scriptscriptstyle\pm}\,9.9$ & $\underline{76.4}^{(2)}\,{\scriptscriptstyle\pm}\,3.4$ & $\underline{95.8}^{(2)}\,{\scriptscriptstyle\pm}\,0.5$ & $\mathbf{69.3}^{(1)}\,{\scriptscriptstyle\pm}\,7.5$ & $\underline{44.8}^{(2)}\,{\scriptscriptstyle\pm}\,2.9$ & $\underline{79.6}^{(2)}\,{\scriptscriptstyle\pm}\,5.0$ & $\mathbf{89.5}^{(1)}\,{\scriptscriptstyle\pm}\,7.3$ & $82.9\,{\scriptscriptstyle\pm}\,6.0$\\
\textbf{WISP-Evolution} & $\underline{74.2}^{(2)}\,{\scriptscriptstyle\pm}\,2.5$ & $\underline{73.0}^{(2)}\,{\scriptscriptstyle\pm}\,6.2$ & $53.3\,{\scriptscriptstyle\pm}\,12.8$ & $\underline{61.4}^{(2)}\,{\scriptscriptstyle\pm}\,9.0$ & $\mathbf{76.7}^{(1)}\,{\scriptscriptstyle\pm}\,3.5$ & $\mathbf{96.2}^{(1)}\,{\scriptscriptstyle\pm}\,0.2$ & $\underline{67.9}^{(2)}\,{\scriptscriptstyle\pm}\,7.9$ & $43.9^{(3)}\,{\scriptscriptstyle\pm}\,3.1$ & $\mathbf{79.8}^{(1)}\,{\scriptscriptstyle\pm}\,4.3$ & $\underline{88.8}^{(2)}\,{\scriptscriptstyle\pm}\,9.3$ & $\mathbf{83.5}^{(1)}\,{\scriptscriptstyle\pm}\,6.0$\\
RDST & $63.9\,{\scriptscriptstyle\pm}\,1.4$ & $59.6\,{\scriptscriptstyle\pm}\,4.0$ & $\mathbf{56.4}^{(1)}\,{\scriptscriptstyle\pm}\,13.5$ & $58.1\,{\scriptscriptstyle\pm}\,5.6$ & $53.9\,{\scriptscriptstyle\pm}\,2.5$ & $79.8\,{\scriptscriptstyle\pm}\,3.2$ & $52.9\,{\scriptscriptstyle\pm}\,5.2$ & $43.5\,{\scriptscriptstyle\pm}\,2.4$ & $68.5^{(3)}\,{\scriptscriptstyle\pm}\,4.8$ & $82.4\,{\scriptscriptstyle\pm}\,6.5$ & $81.2\,{\scriptscriptstyle\pm}\,9.1$\\
MR-Hydra & $62.9\,{\scriptscriptstyle\pm}\,1.3$ & $55.6\,{\scriptscriptstyle\pm}\,3.1$ & $\underline{55.8}^{(2)}\,{\scriptscriptstyle\pm}\,14.0$ & $59.4^{(3)}\,{\scriptscriptstyle\pm}\,7.3$ & $58.3\,{\scriptscriptstyle\pm}\,1.8$ & $79.5\,{\scriptscriptstyle\pm}\,3.5$ & $59.2^{(3)}\,{\scriptscriptstyle\pm}\,6.0$ & $37.6\,{\scriptscriptstyle\pm}\,2.1$ & $67.8\,{\scriptscriptstyle\pm}\,4.4$ & $82.8\,{\scriptscriptstyle\pm}\,4.3$ & $83.1\,{\scriptscriptstyle\pm}\,8.3$\\
MiniRocket & $63.5\,{\scriptscriptstyle\pm}\,1.1$ & $57.3\,{\scriptscriptstyle\pm}\,3.8$ & $52.8\,{\scriptscriptstyle\pm}\,11.0$ & $58.1\,{\scriptscriptstyle\pm}\,6.2$ & $51.7\,{\scriptscriptstyle\pm}\,2.0$ & $77.8\,{\scriptscriptstyle\pm}\,3.8$ & $56.2\,{\scriptscriptstyle\pm}\,5.9$ & $42.4\,{\scriptscriptstyle\pm}\,2.9$ & $68.1\,{\scriptscriptstyle\pm}\,4.8$ & $80.7\,{\scriptscriptstyle\pm}\,6.7$ & $83.5^{(3)}\,{\scriptscriptstyle\pm}\,7.9$\\
MultiRocket & $62.3\,{\scriptscriptstyle\pm}\,1.1$ & $53.8\,{\scriptscriptstyle\pm}\,2.9$ & $55.6^{(3)}\,{\scriptscriptstyle\pm}\,13.4$ & $58.7\,{\scriptscriptstyle\pm}\,7.1$ & $58.3\,{\scriptscriptstyle\pm}\,2.0$ & $79.5\,{\scriptscriptstyle\pm}\,3.5$ & $58.3\,{\scriptscriptstyle\pm}\,6.4$ & $37.1\,{\scriptscriptstyle\pm}\,2.1$ & $67.6\,{\scriptscriptstyle\pm}\,4.4$ & $83.0^{(3)}\,{\scriptscriptstyle\pm}\,4.4$ & $83.3\,{\scriptscriptstyle\pm}\,8.0$\\
Attend-and-Discriminate & $67.8\,{\scriptscriptstyle\pm}\,1.9$ & $64.5\,{\scriptscriptstyle\pm}\,3.9$ & $43.0\,{\scriptscriptstyle\pm}\,8.4$ & $33.0\,{\scriptscriptstyle\pm}\,6.4$ & $52.9\,{\scriptscriptstyle\pm}\,1.9$ & $78.7\,{\scriptscriptstyle\pm}\,2.6$ & $51.2\,{\scriptscriptstyle\pm}\,5.8$ & $\mathbf{46.8}^{(1)}\,{\scriptscriptstyle\pm}\,3.1$ & $65.3\,{\scriptscriptstyle\pm}\,4.6$ & $79.1\,{\scriptscriptstyle\pm}\,2.7$ & $80.8\,{\scriptscriptstyle\pm}\,11.9$\\
H-InceptionTime & $58.1\,{\scriptscriptstyle\pm}\,4.3$ & $57.3\,{\scriptscriptstyle\pm}\,4.1$ & $53.1\,{\scriptscriptstyle\pm}\,8.8$ & $55.0\,{\scriptscriptstyle\pm}\,4.8$ & $52.2\,{\scriptscriptstyle\pm}\,2.2$ & $77.5\,{\scriptscriptstyle\pm}\,6.4$ & $55.6\,{\scriptscriptstyle\pm}\,6.7$ & $39.1\,{\scriptscriptstyle\pm}\,2.3$ & $61.1\,{\scriptscriptstyle\pm}\,5.9$ & $81.1\,{\scriptscriptstyle\pm}\,4.3$ & $82.6\,{\scriptscriptstyle\pm}\,8.9$\\
LITETime & $63.5\,{\scriptscriptstyle\pm}\,1.4$ & $56.3\,{\scriptscriptstyle\pm}\,3.2$ & $49.3\,{\scriptscriptstyle\pm}\,8.7$ & $51.0\,{\scriptscriptstyle\pm}\,5.4$ & $49.9\,{\scriptscriptstyle\pm}\,2.1$ & $76.5\,{\scriptscriptstyle\pm}\,3.7$ & $53.4\,{\scriptscriptstyle\pm}\,7.2$ & $43.1\,{\scriptscriptstyle\pm}\,2.8$ & $62.7\,{\scriptscriptstyle\pm}\,5.1$ & $70.4\,{\scriptscriptstyle\pm}\,7.5$ & $81.8\,{\scriptscriptstyle\pm}\,9.0$\\
QUANT & $69.0^{(3)}\,{\scriptscriptstyle\pm}\,0.7$ & $57.2\,{\scriptscriptstyle\pm}\,3.8$ & $49.8\,{\scriptscriptstyle\pm}\,16.9$ & $54.1\,{\scriptscriptstyle\pm}\,6.6$ & $50.4\,{\scriptscriptstyle\pm}\,1.9$ & $70.0\,{\scriptscriptstyle\pm}\,2.5$ & $50.0\,{\scriptscriptstyle\pm}\,4.8$ & $35.1\,{\scriptscriptstyle\pm}\,2.9$ & $61.4\,{\scriptscriptstyle\pm}\,3.1$ & $69.1\,{\scriptscriptstyle\pm}\,2.7$ & $82.0\,{\scriptscriptstyle\pm}\,4.2$\\
RF-HMM & $68.0\,{\scriptscriptstyle\pm}\,2.0$ & $64.6^{(3)}\,{\scriptscriptstyle\pm}\,8.6$ & $37.5\,{\scriptscriptstyle\pm}\,9.2$ & $4.2\,{\scriptscriptstyle\pm}\,4.0$ & $70.6^{(3)}\,{\scriptscriptstyle\pm}\,2.9$ & $84.9^{(3)}\,{\scriptscriptstyle\pm}\,2.4$ & $45.0\,{\scriptscriptstyle\pm}\,4.2$ & $16.3\,{\scriptscriptstyle\pm}\,1.3$ & $62.3\,{\scriptscriptstyle\pm}\,3.6$ & $68.9\,{\scriptscriptstyle\pm}\,9.1$ & $74.8\,{\scriptscriptstyle\pm}\,6.4$\\
ResNet & $58.2\,{\scriptscriptstyle\pm}\,3.9$ & $59.1\,{\scriptscriptstyle\pm}\,3.8$ & $47.2\,{\scriptscriptstyle\pm}\,10.7$ & $50.5\,{\scriptscriptstyle\pm}\,5.9$ & $51.1\,{\scriptscriptstyle\pm}\,1.6$ & $74.1\,{\scriptscriptstyle\pm}\,5.2$ & $52.5\,{\scriptscriptstyle\pm}\,5.2$ & $38.7\,{\scriptscriptstyle\pm}\,1.1$ & $58.5\,{\scriptscriptstyle\pm}\,3.3$ & $73.3\,{\scriptscriptstyle\pm}\,4.1$ & $81.7\,{\scriptscriptstyle\pm}\,10.1$\\
FCN & $50.8\,{\scriptscriptstyle\pm}\,6.1$ & $55.2\,{\scriptscriptstyle\pm}\,4.1$ & $50.0\,{\scriptscriptstyle\pm}\,11.4$ & $46.3\,{\scriptscriptstyle\pm}\,6.2$ & $50.1\,{\scriptscriptstyle\pm}\,2.3$ & $75.1\,{\scriptscriptstyle\pm}\,3.5$ & $51.6\,{\scriptscriptstyle\pm}\,5.4$ & $39.6\,{\scriptscriptstyle\pm}\,1.7$ & $55.8\,{\scriptscriptstyle\pm}\,4.9$ & $79.3\,{\scriptscriptstyle\pm}\,5.6$ & $\underline{83.5}^{(2)}\,{\scriptscriptstyle\pm}\,9.6$\\
CIF & $67.6\,{\scriptscriptstyle\pm}\,0.5$ & $54.5\,{\scriptscriptstyle\pm}\,3.9$ & $49.0\,{\scriptscriptstyle\pm}\,12.8$ & $47.7\,{\scriptscriptstyle\pm}\,4.9$ & $51.4\,{\scriptscriptstyle\pm}\,2.1$ & $66.4\,{\scriptscriptstyle\pm}\,1.4$ & $39.8\,{\scriptscriptstyle\pm}\,4.3$ & $31.6\,{\scriptscriptstyle\pm}\,1.9$ & $58.1\,{\scriptscriptstyle\pm}\,4.1$ & $70.0\,{\scriptscriptstyle\pm}\,3.9$ & $80.0\,{\scriptscriptstyle\pm}\,1.5$\\
TS2Vec & $63.9\,{\scriptscriptstyle\pm}\,1.6$ & $51.1\,{\scriptscriptstyle\pm}\,3.4$ & $43.2\,{\scriptscriptstyle\pm}\,2.7$ & $44.8\,{\scriptscriptstyle\pm}\,1.2$ & $43.4\,{\scriptscriptstyle\pm}\,2.4$ & $69.7\,{\scriptscriptstyle\pm}\,2.2$ & $43.7\,{\scriptscriptstyle\pm}\,5.4$ & $33.9\,{\scriptscriptstyle\pm}\,2.0$ & $56.4\,{\scriptscriptstyle\pm}\,5.1$ & $59.9\,{\scriptscriptstyle\pm}\,2.7$ & $72.9\,{\scriptscriptstyle\pm}\,8.9$\\
HARNet & $69.0\,{\scriptscriptstyle\pm}\,2.3$ & $54.6\,{\scriptscriptstyle\pm}\,2.6$ & $28.9\,{\scriptscriptstyle\pm}\,4.7$ & $50.8\,{\scriptscriptstyle\pm}\,6.4$ & $42.4\,{\scriptscriptstyle\pm}\,2.1$ & $60.6\,{\scriptscriptstyle\pm}\,2.3$ & $37.4\,{\scriptscriptstyle\pm}\,5.6$ & $32.4\,{\scriptscriptstyle\pm}\,2.5$ & $54.9\,{\scriptscriptstyle\pm}\,5.4$ & $52.7\,{\scriptscriptstyle\pm}\,4.1$ & $58.5\,{\scriptscriptstyle\pm}\,2.6$\\
Balanced-RF & $65.1\,{\scriptscriptstyle\pm}\,0.9$ & $54.6\,{\scriptscriptstyle\pm}\,3.0$ & $38.2\,{\scriptscriptstyle\pm}\,7.7$ & $41.3\,{\scriptscriptstyle\pm}\,4.4$ & $43.5\,{\scriptscriptstyle\pm}\,2.0$ & $62.3\,{\scriptscriptstyle\pm}\,1.5$ & $40.8\,{\scriptscriptstyle\pm}\,4.5$ & $31.0\,{\scriptscriptstyle\pm}\,2.2$ & $49.7\,{\scriptscriptstyle\pm}\,3.2$ & $60.3\,{\scriptscriptstyle\pm}\,4.3$ & $78.3\,{\scriptscriptstyle\pm}\,5.0$\\
PatchTST & $60.4\,{\scriptscriptstyle\pm}\,1.7$ & $57.7\,{\scriptscriptstyle\pm}\,3.5$ & $30.0\,{\scriptscriptstyle\pm}\,10.4$ & $44.4\,{\scriptscriptstyle\pm}\,3.7$ & $38.4\,{\scriptscriptstyle\pm}\,1.0$ & $61.0\,{\scriptscriptstyle\pm}\,1.2$ & $32.0\,{\scriptscriptstyle\pm}\,3.5$ & $28.9\,{\scriptscriptstyle\pm}\,1.5$ & $45.6\,{\scriptscriptstyle\pm}\,4.7$ & $47.7\,{\scriptscriptstyle\pm}\,2.4$ & $66.3\,{\scriptscriptstyle\pm}\,3.1$\\
TimesNet & $56.7\,{\scriptscriptstyle\pm}\,1.9$ & $52.9\,{\scriptscriptstyle\pm}\,5.6$ & $33.1\,{\scriptscriptstyle\pm}\,8.3$ & $39.8\,{\scriptscriptstyle\pm}\,2.5$ & $38.5\,{\scriptscriptstyle\pm}\,1.4$ & $61.9\,{\scriptscriptstyle\pm}\,2.4$ & $28.8\,{\scriptscriptstyle\pm}\,1.8$ & $29.1\,{\scriptscriptstyle\pm}\,1.1$ & $46.3\,{\scriptscriptstyle\pm}\,4.3$ & $53.5\,{\scriptscriptstyle\pm}\,1.7$ & $60.0\,{\scriptscriptstyle\pm}\,7.6$\\
TF-C & $53.5\,{\scriptscriptstyle\pm}\,1.1$ & $33.6\,{\scriptscriptstyle\pm}\,2.0$ & $24.1\,{\scriptscriptstyle\pm}\,4.4$ & $38.9\,{\scriptscriptstyle\pm}\,3.9$ & $22.7\,{\scriptscriptstyle\pm}\,1.2$ & $46.1\,{\scriptscriptstyle\pm}\,1.9$ & $33.9\,{\scriptscriptstyle\pm}\,4.2$ & $17.5\,{\scriptscriptstyle\pm}\,1.0$ & $41.6\,{\scriptscriptstyle\pm}\,3.4$ & $41.9\,{\scriptscriptstyle\pm}\,4.2$ & $53.5\,{\scriptscriptstyle\pm}\,4.0$\\
\bottomrule\end{tabularx}
\end{table}
\end{landscape}
\clearpage

\begin{table}[H]
\centering\fontsize{7.8}{9.2}\selectfont\setlength{\tabcolsep}{0pt}\renewcommand{\arraystretch}{0.95}
\caption{External comparison: accuracy percentages (mean of five split means; standard deviations are in the released results). Methods, starred tasks and rank markings follow Table~\ref{tab:rev6_external_macro_f1}.}\label{tab:rev6_external_accuracy}
\begin{tabular*}{\linewidth}{@{\extracolsep{\fill}}l*{11}{r}@{}}
\toprule Method & \rotatebox{90}{CAPTURE-24} & \rotatebox{90}{WISDM} & \rotatebox{90}{ADL} & \rotatebox{90}{IUWDS} & \rotatebox{90}{PAAL} & \rotatebox{90}{HANDY} & \rotatebox{90}{DOMINO} & \rotatebox{90}{HARMES} & \rotatebox{90}{GOTOV} & \rotatebox{90}{PAMAP2*} & \rotatebox{90}{MHEALTH*}\\\midrule
\textbf{WISP-Random} & $\mathbf{83.5}^{(1)}$ & $\mathbf{74.2}^{(1)}$ & $65.6$ & $\mathbf{85.2}^{(1)}$ & $76.6^{(3)}$ & $\underline{94.9}^{(2)}$ & $\mathbf{86.5}^{(1)}$ & $48.4$ & $\underline{80.6}^{(2)}$ & $\mathbf{90.0}^{(1)}$ & $83.0$\\
\textbf{WISP-Evolution} & $\underline{82.8}^{(2)}$ & $\underline{73.1}^{(2)}$ & $65.6$ & $\underline{83.8}^{(2)}$ & $\underline{77.0}^{(2)}$ & $\mathbf{95.5}^{(1)}$ & $\underline{85.5}^{(2)}$ & $47.3$ & $\mathbf{80.9}^{(1)}$ & $\underline{89.2}^{(2)}$ & $\underline{84.0}^{(2)}$\\
RDST & $76.2$ & $59.3$ & $\mathbf{70.4}^{(1)}$ & $78.7$ & $63.4$ & $81.4$ & $77.8$ & $48.8^{(3)}$ & $70.4$ & $82.7$ & $80.9$\\
MR-Hydra & $72.7$ & $54.7$ & $\underline{68.6}^{(2)}$ & $78.0$ & $65.5$ & $81.3$ & $79.6^{(3)}$ & $41.7$ & $69.3$ & $83.4$ & $82.9$\\
MiniRocket & $76.7$ & $57.5$ & $65.8$ & $77.8$ & $62.1$ & $79.6$ & $77.9$ & $\underline{49.2}^{(2)}$ & $71.1^{(3)}$ & $81.3$ & $83.3^{(3)}$\\
MultiRocket & $72.0$ & $52.9$ & $68.3^{(3)}$ & $77.4$ & $65.2$ & $81.1$ & $79.0$ & $41.1$ & $69.0$ & $83.2$ & $83.1$\\
Attend-and-Discriminate & $74.9$ & $64.4$ & $53.7$ & $53.7$ & $60.9$ & $80.7$ & $59.8$ & $\mathbf{49.5}^{(1)}$ & $69.1$ & $83.5^{(3)}$ & $81.3$\\
H-InceptionTime & $67.8$ & $57.2$ & $68.3$ & $75.1$ & $61.2$ & $79.9$ & $74.2$ & $41.7$ & $64.5$ & $81.8$ & $82.8$\\
LITETime & $73.2$ & $56.1$ & $64.9$ & $73.1$ & $58.3$ & $78.8$ & $70.3$ & $46.8$ & $67.3$ & $74.3$ & $82.2$\\
QUANT & $77.9$ & $57.6$ & $63.8$ & $79.2^{(3)}$ & $59.9$ & $73.2$ & $73.9$ & $41.6$ & $65.0$ & $73.2$ & $81.9$\\
RF-HMM & $78.9^{(3)}$ & $65.9^{(3)}$ & $54.1$ & $6.0$ & $\mathbf{79.0}^{(1)}$ & $86.4^{(3)}$ & $62.5$ & $22.6$ & $65.6$ & $74.1$ & $74.7$\\
ResNet & $67.6$ & $58.8$ & $63.3$ & $72.9$ & $60.0$ & $76.4$ & $72.5$ & $41.4$ & $62.6$ & $76.7$ & $82.2$\\
FCN & $61.6$ & $54.7$ & $62.6$ & $71.1$ & $57.9$ & $76.1$ & $68.4$ & $42.6$ & $60.2$ & $82.1$ & $\mathbf{84.0}^{(1)}$\\
CIF & $77.7$ & $55.3$ & $63.9$ & $78.7$ & $59.6$ & $71.0$ & $70.1$ & $39.2$ & $66.2$ & $71.8$ & $80.0$\\
TS2Vec & $71.2$ & $51.6$ & $58.8$ & $67.1$ & $52.1$ & $71.5$ & $53.8$ & $36.6$ & $62.5$ & $65.4$ & $73.3$\\
HARNet & $75.5$ & $54.9$ & $43.4$ & $71.3$ & $50.4$ & $62.2$ & $46.5$ & $35.1$ & $58.0$ & $56.0$ & $58.6$\\
Balanced-RF & $76.0$ & $55.1$ & $56.1$ & $69.9$ & $52.8$ & $67.0$ & $69.4$ & $36.8$ & $57.5$ & $66.1$ & $79.4$\\
PatchTST & $67.6$ & $58.0$ & $39.7$ & $62.4$ & $45.5$ & $64.0$ & $38.3$ & $30.8$ & $52.3$ & $51.4$ & $68.1$\\
TimesNet & $64.6$ & $52.5$ & $43.4$ & $61.9$ & $46.3$ & $64.4$ & $38.7$ & $31.2$ & $53.0$ & $58.5$ & $61.8$\\
TF-C & $60.3$ & $34.7$ & $35.9$ & $59.7$ & $27.5$ & $47.7$ & $40.5$ & $19.6$ & $46.3$ & $43.5$ & $52.9$\\
\bottomrule\end{tabular*}
\end{table}

\begin{table}[H]
\centering\fontsize{7.8}{9.2}\selectfont\setlength{\tabcolsep}{0pt}\renewcommand{\arraystretch}{0.95}
\caption{External comparison: balanced accuracy percentages (mean of five split means; standard deviations are in the released results). Methods, starred tasks and rank markings follow Table~\ref{tab:rev6_external_macro_f1}.}\label{tab:rev6_external_balanced_accuracy}
\begin{tabular*}{\linewidth}{@{\extracolsep{\fill}}l*{11}{r}@{}}
\toprule Method & \rotatebox{90}{CAPTURE-24} & \rotatebox{90}{WISDM} & \rotatebox{90}{ADL} & \rotatebox{90}{IUWDS} & \rotatebox{90}{PAAL} & \rotatebox{90}{HANDY} & \rotatebox{90}{DOMINO} & \rotatebox{90}{HARMES} & \rotatebox{90}{GOTOV} & \rotatebox{90}{PAMAP2*} & \rotatebox{90}{MHEALTH*}\\\midrule
\textbf{WISP-Random} & $\mathbf{73.4}^{(1)}$ & $\mathbf{74.3}^{(1)}$ & $57.7$ & $\mathbf{62.7}^{(1)}$ & $\underline{75.0}^{(2)}$ & $\underline{96.1}^{(2)}$ & $\mathbf{70.8}^{(1)}$ & $47.1^{(3)}$ & $\mathbf{80.8}^{(1)}$ & $\mathbf{88.9}^{(1)}$ & $83.4$\\
\textbf{WISP-Evolution} & $\underline{73.4}^{(2)}$ & $\underline{73.1}^{(2)}$ & $56.1$ & $\underline{60.1}^{(2)}$ & $\mathbf{75.3}^{(1)}$ & $\mathbf{96.5}^{(1)}$ & $\underline{68.9}^{(2)}$ & $\underline{47.2}^{(2)}$ & $\underline{80.8}^{(2)}$ & $\underline{88.3}^{(2)}$ & $\mathbf{84.2}^{(1)}$\\
RDST & $62.4$ & $59.4$ & $\mathbf{64.0}^{(1)}$ & $54.6$ & $53.9$ & $78.8$ & $53.8$ & $42.3$ & $67.6$ & $82.1$ & $81.9$\\
MR-Hydra & $61.4$ & $54.8$ & $\underline{63.5}^{(2)}$ & $56.8$ & $57.6$ & $78.8$ & $60.3$ & $36.4$ & $66.7$ & $83.0$ & $83.7$\\
MiniRocket & $62.0$ & $57.6$ & $61.4$ & $55.3$ & $51.8$ & $77.1$ & $56.9$ & $41.5$ & $67.4$ & $80.3$ & $\underline{84.0}^{(2)}$\\
MultiRocket & $60.8$ & $52.9$ & $63.4$ & $55.7$ & $57.6$ & $78.7$ & $59.8$ & $36.0$ & $66.4$ & $82.9$ & $83.8$\\
Attend-and-Discriminate & $70.7$ & $64.5$ & $57.4$ & $47.4$ & $55.6$ & $79.8$ & $63.2^{(3)}$ & $\mathbf{50.0}^{(1)}$ & $70.6^{(3)}$ & $83.9^{(3)}$ & $82.2$\\
H-InceptionTime & $59.1$ & $57.3$ & $63.5^{(3)}$ & $56.0$ & $53.3$ & $77.7$ & $57.7$ & $39.5$ & $62.6$ & $81.5$ & $83.3$\\
LITETime & $62.7$ & $56.2$ & $60.9$ & $53.7$ & $50.9$ & $76.5$ & $58.3$ & $43.0$ & $64.2$ & $74.4$ & $82.7$\\
QUANT & $66.7$ & $57.7$ & $55.2$ & $50.9$ & $51.1$ & $69.1$ & $51.0$ & $33.7$ & $61.0$ & $71.3$ & $82.7$\\
RF-HMM & $67.5$ & $65.9^{(3)}$ & $46.6$ & $19.2$ & $71.2^{(3)}$ & $84.6^{(3)}$ & $53.4$ & $21.2$ & $67.0$ & $72.6$ & $75.8$\\
ResNet & $59.0$ & $58.9$ & $58.1$ & $51.3$ & $51.6$ & $74.8$ & $54.9$ & $38.9$ & $59.4$ & $75.9$ & $82.2$\\
FCN & $53.5$ & $54.8$ & $59.7$ & $49.1$ & $51.7$ & $76.2$ & $54.0$ & $40.4$ & $57.0$ & $81.9$ & $83.9^{(3)}$\\
CIF & $65.4$ & $55.4$ & $55.3$ & $45.8$ & $51.8$ & $65.5$ & $42.5$ & $30.7$ & $58.1$ & $70.1$ & $80.5$\\
TS2Vec & $68.7$ & $51.6$ & $55.9$ & $54.9$ & $45.2$ & $70.6$ & $52.4$ & $37.0$ & $60.0$ & $63.8$ & $73.2$\\
HARNet & $71.4^{(3)}$ & $54.9$ & $39.0$ & $58.3^{(3)}$ & $44.9$ & $62.0$ & $46.6$ & $34.4$ & $58.2$ & $55.7$ & $59.3$\\
Balanced-RF & $62.7$ & $55.2$ & $46.2$ & $39.6$ & $44.4$ & $62.2$ & $42.9$ & $29.8$ & $49.9$ & $62.8$ & $79.6$\\
PatchTST & $65.1$ & $58.0$ & $42.4$ & $54.5$ & $41.0$ & $61.9$ & $39.6$ & $31.4$ & $49.1$ & $51.5$ & $66.9$\\
TimesNet & $60.9$ & $52.6$ & $45.0$ & $47.7$ & $40.1$ & $62.7$ & $34.7$ & $31.8$ & $49.7$ & $57.3$ & $61.2$\\
TF-C & $56.8$ & $34.7$ & $33.7$ & $44.8$ & $25.9$ & $46.8$ & $42.1$ & $20.5$ & $43.5$ & $44.4$ & $54.6$\\
\bottomrule\end{tabular*}
\end{table}

\begin{table}[H]
\centering\fontsize{7.8}{9.2}\selectfont\setlength{\tabcolsep}{0pt}\renewcommand{\arraystretch}{0.95}
\caption{External comparison: worst-class F1 percentages (mean of five split means; standard deviations are in the released results). Methods, starred tasks and rank markings follow Table~\ref{tab:rev6_external_macro_f1}.}\label{tab:rev6_external_worst_class_f1}
\begin{tabular*}{\linewidth}{@{\extracolsep{\fill}}l*{11}{r}@{}}
\toprule Method & \rotatebox{90}{CAPTURE-24} & \rotatebox{90}{WISDM} & \rotatebox{90}{ADL} & \rotatebox{90}{IUWDS} & \rotatebox{90}{PAAL} & \rotatebox{90}{HANDY} & \rotatebox{90}{DOMINO} & \rotatebox{90}{HARMES} & \rotatebox{90}{GOTOV} & \rotatebox{90}{PAMAP2*} & \rotatebox{90}{MHEALTH*}\\\midrule
\textbf{WISP-Random} & $\mathbf{51.9}^{(1)}$ & $\mathbf{28.3}^{(1)}$ & $0.0$ & $0.0^{(3)}$ & $\mathbf{43.3}^{(1)}$ & $\underline{91.1}^{(2)}$ & $0.0$ & $\underline{18.7}^{(2)}$ & $21.0$ & $\underline{27.0}^{(2)}$ & $46.1$\\
\textbf{WISP-Evolution} & $\underline{51.8}^{(2)}$ & $25.8^{(3)}$ & $0.0$ & $0.0^{(3)}$ & $\underline{42.6}^{(2)}$ & $\mathbf{92.1}^{(1)}$ & $0.0$ & $18.3^{(3)}$ & $21.0$ & $\mathbf{27.0}^{(1)}$ & $42.5$\\
RDST & $29.8$ & $21.4$ & $0.3$ & $0.0^{(3)}$ & $8.5$ & $61.4$ & $0.0$ & $5.5$ & $28.5$ & $10.5$ & $40.4$\\
MR-Hydra & $37.6$ & $19.2$ & $0.3$ & $0.0^{(3)}$ & $15.8$ & $57.5$ & $0.0$ & $8.2$ & $\underline{30.9}^{(2)}$ & $22.1^{(3)}$ & $45.2$\\
MiniRocket & $27.1$ & $18.3$ & $1.0$ & $0.0^{(3)}$ & $6.3$ & $56.0$ & $0.0$ & $0.5$ & $29.0^{(3)}$ & $11.9$ & $\underline{46.9}^{(2)}$\\
MultiRocket & $37.1$ & $18.5$ & $0.3$ & $0.0^{(3)}$ & $17.8$ & $57.5$ & $0.3$ & $8.5$ & $\mathbf{31.0}^{(1)}$ & $20.4$ & $46.2^{(3)}$\\
Attend-and-Discriminate & $47.8$ & $\underline{26.9}^{(2)}$ & $1.3$ & $\underline{0.8}^{(2)}$ & $18.1$ & $62.5$ & $\mathbf{3.4}^{(1)}$ & $\mathbf{21.4}^{(1)}$ & $22.5$ & $22.1$ & $43.3$\\
H-InceptionTime & $32.5$ & $20.6$ & $2.4^{(3)}$ & $0.0^{(3)}$ & $15.0$ & $56.0$ & $1.7$ & $14.5$ & $19.6$ & $21.2$ & $38.2$\\
LITETime & $36.9$ & $19.6$ & $\underline{2.6}^{(2)}$ & $0.0^{(3)}$ & $12.4$ & $54.3$ & $2.3$ & $13.4$ & $22.3$ & $8.8$ & $41.0$\\
QUANT & $46.5$ & $13.8$ & $0.0$ & $0.0^{(3)}$ & $7.9$ & $40.3$ & $0.0$ & $7.8$ & $24.0$ & $14.6$ & $\mathbf{52.8}^{(1)}$\\
RF-HMM & $44.2$ & $8.9$ & $0.0$ & $0.0^{(3)}$ & $24.7^{(3)}$ & $68.0^{(3)}$ & $0.0$ & $0.0$ & $13.1$ & $0.0$ & $25.0$\\
ResNet & $31.5$ & $22.7$ & $1.2$ & $0.0^{(3)}$ & $14.7$ & $52.6$ & $0.6$ & $11.7$ & $18.6$ & $16.9$ & $43.5$\\
FCN & $25.1$ & $19.4$ & $0.9$ & $0.0^{(3)}$ & $11.5$ & $48.9$ & $1.1$ & $9.1$ & $9.0$ & $20.8$ & $41.5$\\
CIF & $41.3$ & $13.8$ & $0.0$ & $0.0^{(3)}$ & $9.7$ & $32.8$ & $0.0$ & $1.9$ & $0.0$ & $12.0$ & $42.7$\\
TS2Vec & $42.2$ & $15.2$ & $\mathbf{3.1}^{(1)}$ & $0.0^{(3)}$ & $10.2$ & $51.4$ & $2.4^{(3)}$ & $12.2$ & $18.7$ & $10.9$ & $35.7$\\
HARNet & $50.6^{(3)}$ & $19.6$ & $0.9$ & $0.0^{(3)}$ & $9.3$ & $40.3$ & $\underline{2.6}^{(2)}$ & $11.9$ & $21.9$ & $9.2$ & $22.8$\\
Balanced-RF & $37.7$ & $13.4$ & $0.0$ & $0.0^{(3)}$ & $3.7$ & $20.2$ & $0.0$ & $5.8$ & $0.0$ & $9.2$ & $40.7$\\
PatchTST & $37.9$ & $21.4$ & $0.1$ & $\mathbf{2.4}^{(1)}$ & $8.8$ & $39.2$ & $2.1$ & $8.0$ & $8.2$ & $7.8$ & $32.4$\\
TimesNet & $32.0$ & $19.3$ & $0.5$ & $0.0^{(3)}$ & $7.9$ & $40.0$ & $0.9$ & $7.1$ & $7.9$ & $7.6$ & $18.8$\\
TF-C & $34.6$ & $7.2$ & $0.7$ & $0.0^{(3)}$ & $1.4$ & $21.7$ & $2.4$ & $4.7$ & $6.2$ & $8.7$ & $25.0$\\
\bottomrule\end{tabular*}
\end{table}

\ifdefined\ExtendedMode
\else
\wispinternalheading

\FloatBarrier
\fi

\subsection{Search budget and trajectories}\label{app:budget_curves}\label{app:search_concentration}

Table~\ref{tab:budget_tests} reports matched-budget effects, and Figure~\ref{fig:grid_macro_f1} shows per-dataset trajectories. On HARMES, $H$ is disabled in 30/30 Random and 28/30 Evolution selected configurations (two wrists, five splits and three seeds).

\begin{table}[H]\centering\footnotesize\setlength{\tabcolsep}{4pt}\renewcommand{\arraystretch}{0.90}
\caption{Evolution minus Random at each candidate budget and for normalised log-budget area under the curve (AUC), in percentage points over nine dataset means. AUC uses trapezoidal integration over log budget, divided by the log-budget range. Exact sign-flip tests are Holm-adjusted across the five budgets within each measure; AUC rows report the unadjusted exact test.}\label{tab:budget_tests}
\begin{tabularx}{\linewidth}{@{}Xlrrrc@{}}
\toprule Measure & Budget & Change & 95\% interval & $p$ & W/T/L\\\midrule
Macro-F1 & 25 & +0.49 & [-0.68, +1.86] & 0.758 & 5/0/4\\
 & 50 & +0.47 & [-0.23, +1.30] & 0.758 & 4/0/5\\
 & 100 & -0.38 & [-0.91, +0.15] & 0.680 & 4/0/5\\
 & 200 & -0.53 & [-1.05, -0.06] & 0.328 & 2/0/7\\
 & 329 & -0.61 & [-1.10, -0.12] & 0.273 & 3/0/6\\
\addlinespace[2pt]
 & AUC & -0.09 & [-0.52, +0.31] & 0.691 & 5/0/4\\\addlinespace[3pt]
Accuracy & 25 & +0.74 & [+0.03, +1.79] & 0.430 & 6/0/3\\
 & 50 & +0.42 & [-0.24, +1.26] & 0.812 & 5/0/4\\
 & 100 & -0.21 & [-0.79, +0.31] & 0.812 & 5/0/4\\
 & 200 & -0.43 & [-0.81, -0.01] & 0.430 & 2/0/7\\
 & 329 & -0.44 & [-0.90, +0.04] & 0.430 & 4/0/5\\
\addlinespace[2pt]
 & AUC & +0.01 & [-0.45, +0.51] & 0.973 & 5/0/4\\\addlinespace[3pt]
Balanced accuracy & 25 & +0.37 & [-1.04, +1.84] & 0.680 & 5/0/4\\
 & 50 & +0.38 & [-0.28, +1.09] & 0.680 & 5/0/4\\
 & 100 & -0.53 & [-1.30, +0.16] & 0.680 & 5/0/4\\
 & 200 & -0.77 & [-1.45, -0.17] & 0.215 & 2/0/7\\
 & 329 & -0.71 & [-1.40, -0.06] & 0.438 & 3/0/6\\
\addlinespace[2pt]
 & AUC & -0.24 & [-0.85, +0.23] & 0.547 & 6/0/3\\\addlinespace[3pt]
Worst-class F1 & 25 & -0.04 & [-0.95, +0.78] & 1.000 & 3/3/3\\
 & 50 & +0.29 & [-0.15, +0.86] & 1.000 & 4/3/2\\
 & 100 & -0.61 & [-1.93, +0.49] & 1.000 & 3/3/3\\
 & 200 & -0.15 & [-0.64, +0.30] & 1.000 & 3/3/3\\
 & 329 & -0.29 & [-0.92, +0.22] & 1.000 & 1/3/5\\
\addlinespace[2pt]
 & AUC & -0.15 & [-0.56, +0.14] & 0.625 & 4/3/2\\\addlinespace[3pt]
\bottomrule\end{tabularx}\end{table}

\begin{figure}[htbp]
\centering\includegraphics[width=\linewidth]{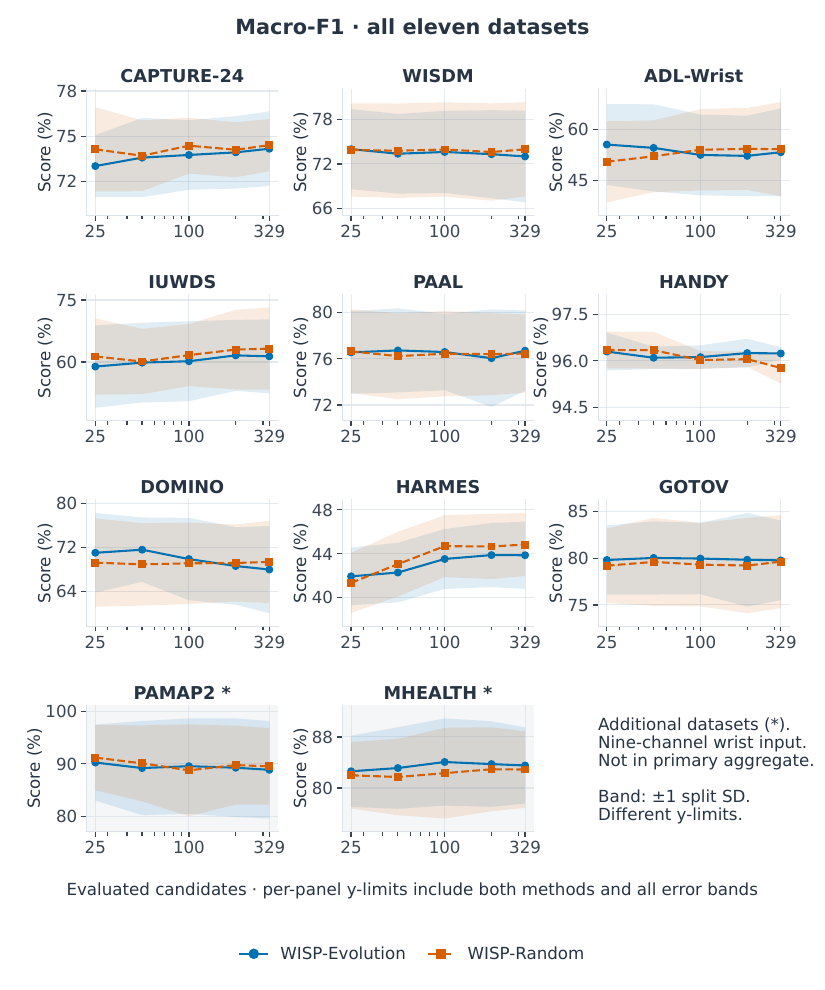}
\caption{Per-dataset Macro-F1 versus candidate budget for validation-selected configurations evaluated on test data: WISP-Evolution (blue circles) and WISP-Random (orange squares). Bands show one standard deviation across five seed-averaged participant splits, pairing HARMES wrists within split. Panels use separate vertical ranges; the first nine datasets are primary and the two starred tasks are additional analyses.}
\label{fig:grid_macro_f1}
\end{figure}
\FloatBarrier

\begin{landscape}
\begin{table}[H]
\centering\fontsize{8.8}{10.0}\selectfont\setlength{\tabcolsep}{1.2pt}\renewcommand{\arraystretch}{1.0}
\ifdefined\ExtendedMode\else
\wispinternalmatrixscope\par
\fi
\caption{Internal comparison: Macro-F1 percentages (mean $\pm$ standard deviation across five split means). Eleven complete designs and fourteen WISP-CIS recipes are ranked separately; $H$ denotes Viterbi decoding. Starred tasks and rank markings follow Table~\ref{tab:rev6_external_macro_f1}.}\label{tab:rev6_internal_macro_f1}
\begin{tabularx}{\linewidth}{@{}>{\raggedright\arraybackslash}Xrrrrrrrrrrr@{}}
\toprule Method / recipe & CAPTURE-24 & WISDM & ADL & IUWDS & PAAL & HANDY & DOMINO & HARMES & GOTOV & PAMAP2* & MHEALTH*\\\midrule
\multicolumn{12}{@{}l}{\textit{Complete WISP designs}}\\
WISP-SS & $59.8\,{\scriptscriptstyle\pm}\,2.2$ & $74.8\,{\scriptscriptstyle\pm}\,5.2$ & $40.9\,{\scriptscriptstyle\pm}\,13.9$ & $41.0\,{\scriptscriptstyle\pm}\,5.6$ & $75.6\,{\scriptscriptstyle\pm}\,4.0$ & $\mathbf{96.4}^{(1)}\,{\scriptscriptstyle\pm}\,0.5$ & $45.2\,{\scriptscriptstyle\pm}\,3.0$ & $21.6\,{\scriptscriptstyle\pm}\,1.8$ & $71.5\,{\scriptscriptstyle\pm}\,3.2$ & $86.9\,{\scriptscriptstyle\pm}\,3.5$ & $87.8^{(3)}\,{\scriptscriptstyle\pm}\,7.8$\\
WISP-RC & $71.9\,{\scriptscriptstyle\pm}\,1.9$ & $72.3\,{\scriptscriptstyle\pm}\,6.6$ & $48.0\,{\scriptscriptstyle\pm}\,10.8$ & $57.2\,{\scriptscriptstyle\pm}\,9.3$ & $60.2\,{\scriptscriptstyle\pm}\,2.6$ & $88.5\,{\scriptscriptstyle\pm}\,0.8$ & $\mathbf{72.4}^{(1)}\,{\scriptscriptstyle\pm}\,4.9$ & $21.5\,{\scriptscriptstyle\pm}\,1.0$ & $79.2\,{\scriptscriptstyle\pm}\,4.7$ & $\underline{95.0}^{(2)}\,{\scriptscriptstyle\pm}\,1.7$ & $\underline{87.9}^{(2)}\,{\scriptscriptstyle\pm}\,7.2$\\
WISP-SO & $55.4\,{\scriptscriptstyle\pm}\,2.5$ & $71.6\,{\scriptscriptstyle\pm}\,5.8$ & $41.7\,{\scriptscriptstyle\pm}\,17.0$ & $39.6\,{\scriptscriptstyle\pm}\,6.1$ & $67.6\,{\scriptscriptstyle\pm}\,4.1$ & $94.7\,{\scriptscriptstyle\pm}\,0.4$ & $33.8\,{\scriptscriptstyle\pm}\,6.5$ & $14.3\,{\scriptscriptstyle\pm}\,1.4$ & $63.6\,{\scriptscriptstyle\pm}\,4.4$ & $75.1\,{\scriptscriptstyle\pm}\,8.4$ & $85.8\,{\scriptscriptstyle\pm}\,9.2$\\
WISP-CSE & $59.9\,{\scriptscriptstyle\pm}\,2.2$ & $74.9\,{\scriptscriptstyle\pm}\,5.8$ & $42.4\,{\scriptscriptstyle\pm}\,16.2$ & $41.7\,{\scriptscriptstyle\pm}\,5.3$ & $75.9^{(3)}\,{\scriptscriptstyle\pm}\,4.4$ & $95.0\,{\scriptscriptstyle\pm}\,0.4$ & $46.2\,{\scriptscriptstyle\pm}\,3.8$ & $18.8\,{\scriptscriptstyle\pm}\,1.8$ & $73.9\,{\scriptscriptstyle\pm}\,4.8$ & $89.5\,{\scriptscriptstyle\pm}\,2.6$ & $\mathbf{88.8}^{(1)}\,{\scriptscriptstyle\pm}\,7.0$\\
WISP-GIS & $49.2\,{\scriptscriptstyle\pm}\,2.5$ & $72.7\,{\scriptscriptstyle\pm}\,5.6$ & $41.8\,{\scriptscriptstyle\pm}\,13.1$ & $40.8\,{\scriptscriptstyle\pm}\,5.1$ & $69.4\,{\scriptscriptstyle\pm}\,5.2$ & $94.8\,{\scriptscriptstyle\pm}\,0.5$ & $39.9\,{\scriptscriptstyle\pm}\,4.9$ & $15.8\,{\scriptscriptstyle\pm}\,1.1$ & $72.0\,{\scriptscriptstyle\pm}\,4.6$ & $70.6\,{\scriptscriptstyle\pm}\,7.3$ & $83.4\,{\scriptscriptstyle\pm}\,5.2$\\
WISP-CIS & $74.0^{(3)}\,{\scriptscriptstyle\pm}\,1.9$ & $\underline{75.8}^{(2)}\,{\scriptscriptstyle\pm}\,4.9$ & $51.2^{(3)}\,{\scriptscriptstyle\pm}\,12.2$ & $59.4^{(3)}\,{\scriptscriptstyle\pm}\,9.2$ & $72.0\,{\scriptscriptstyle\pm}\,3.6$ & $89.6\,{\scriptscriptstyle\pm}\,0.7$ & $68.8\,{\scriptscriptstyle\pm}\,7.4$ & $24.0^{(3)}\,{\scriptscriptstyle\pm}\,1.6$ & $\mathbf{80.4}^{(1)}\,{\scriptscriptstyle\pm}\,3.5$ & $\mathbf{95.8}^{(1)}\,{\scriptscriptstyle\pm}\,1.4$ & $85.8\,{\scriptscriptstyle\pm}\,8.5$\\
WISP-ESE & $56.7\,{\scriptscriptstyle\pm}\,2.5$ & $73.9\,{\scriptscriptstyle\pm}\,5.6$ & $43.5\,{\scriptscriptstyle\pm}\,15.8$ & $42.7\,{\scriptscriptstyle\pm}\,5.1$ & $72.6\,{\scriptscriptstyle\pm}\,4.6$ & $94.3\,{\scriptscriptstyle\pm}\,0.4$ & $43.4\,{\scriptscriptstyle\pm}\,5.2$ & $16.1\,{\scriptscriptstyle\pm}\,1.6$ & $73.2\,{\scriptscriptstyle\pm}\,4.6$ & $83.1\,{\scriptscriptstyle\pm}\,4.9$ & $86.3\,{\scriptscriptstyle\pm}\,8.3$\\
WISP-Select5 & $74.0^{(3)}\,{\scriptscriptstyle\pm}\,1.9$ & $\mathbf{75.8}^{(1)}\,{\scriptscriptstyle\pm}\,5.0$ & $50.0\,{\scriptscriptstyle\pm}\,11.5$ & $59.4^{(3)}\,{\scriptscriptstyle\pm}\,9.2$ & $74.7\,{\scriptscriptstyle\pm}\,4.4$ & $\underline{96.3}^{(2)}\,{\scriptscriptstyle\pm}\,0.6$ & $\underline{70.4}^{(2)}\,{\scriptscriptstyle\pm}\,7.4$ & $23.5\,{\scriptscriptstyle\pm}\,1.3$ & $79.7^{(3)}\,{\scriptscriptstyle\pm}\,4.2$ & $91.6^{(3)}\,{\scriptscriptstyle\pm}\,5.6$ & $86.2\,{\scriptscriptstyle\pm}\,7.6$\\
WISP-Select7 & $74.0^{(3)}\,{\scriptscriptstyle\pm}\,1.9$ & $75.4^{(3)}\,{\scriptscriptstyle\pm}\,5.5$ & $46.9\,{\scriptscriptstyle\pm}\,10.4$ & $59.4^{(3)}\,{\scriptscriptstyle\pm}\,9.2$ & $75.1\,{\scriptscriptstyle\pm}\,3.9$ & $96.0\,{\scriptscriptstyle\pm}\,0.7$ & $\underline{70.4}^{(2)}\,{\scriptscriptstyle\pm}\,7.4$ & $23.5\,{\scriptscriptstyle\pm}\,1.3$ & $79.5\,{\scriptscriptstyle\pm}\,4.6$ & $91.6^{(3)}\,{\scriptscriptstyle\pm}\,5.6$ & $86.2\,{\scriptscriptstyle\pm}\,7.6$\\
\textbf{WISP-Evolution} & $\underline{74.2}^{(2)}\,{\scriptscriptstyle\pm}\,2.5$ & $73.0\,{\scriptscriptstyle\pm}\,6.2$ & $\underline{53.3}^{(2)}\,{\scriptscriptstyle\pm}\,12.8$ & $\underline{61.4}^{(2)}\,{\scriptscriptstyle\pm}\,9.0$ & $\mathbf{76.7}^{(1)}\,{\scriptscriptstyle\pm}\,3.5$ & $96.2^{(3)}\,{\scriptscriptstyle\pm}\,0.2$ & $67.9\,{\scriptscriptstyle\pm}\,7.9$ & $\underline{43.9}^{(2)}\,{\scriptscriptstyle\pm}\,3.1$ & $\underline{79.8}^{(2)}\,{\scriptscriptstyle\pm}\,4.3$ & $88.8\,{\scriptscriptstyle\pm}\,9.3$ & $83.5\,{\scriptscriptstyle\pm}\,6.0$\\
\textbf{WISP-Random} & $\mathbf{74.4}^{(1)}\,{\scriptscriptstyle\pm}\,1.7$ & $74.0\,{\scriptscriptstyle\pm}\,6.3$ & $\mathbf{54.2}^{(1)}\,{\scriptscriptstyle\pm}\,13.7$ & $\mathbf{63.2}^{(1)}\,{\scriptscriptstyle\pm}\,9.9$ & $\underline{76.4}^{(2)}\,{\scriptscriptstyle\pm}\,3.4$ & $95.8\,{\scriptscriptstyle\pm}\,0.5$ & $69.3\,{\scriptscriptstyle\pm}\,7.5$ & $\mathbf{44.8}^{(1)}\,{\scriptscriptstyle\pm}\,2.9$ & $79.6\,{\scriptscriptstyle\pm}\,5.0$ & $89.5\,{\scriptscriptstyle\pm}\,7.3$ & $82.9\,{\scriptscriptstyle\pm}\,6.0$\\
\addlinespace[2pt]
\multicolumn{12}{@{}l}{\textit{Controlled WISP-CIS recipes}}\\
R + direct & $65.9\,{\scriptscriptstyle\pm}\,1.9$ & $56.8\,{\scriptscriptstyle\pm}\,4.3$ & $48.8\,{\scriptscriptstyle\pm}\,7.0$ & $52.5\,{\scriptscriptstyle\pm}\,2.3$ & $51.9\,{\scriptscriptstyle\pm}\,1.8$ & $77.2\,{\scriptscriptstyle\pm}\,2.7$ & $56.4\,{\scriptscriptstyle\pm}\,7.0$ & $41.7^{(3)}\,{\scriptscriptstyle\pm}\,2.7$ & $64.3\,{\scriptscriptstyle\pm}\,3.7$ & $82.1\,{\scriptscriptstyle\pm}\,3.2$ & $82.4\,{\scriptscriptstyle\pm}\,7.2$\\
R + H & $71.9\,{\scriptscriptstyle\pm}\,1.9$ & $72.3\,{\scriptscriptstyle\pm}\,6.6$ & $48.0\,{\scriptscriptstyle\pm}\,10.8$ & $57.2\,{\scriptscriptstyle\pm}\,9.3$ & $60.2\,{\scriptscriptstyle\pm}\,2.6$ & $88.5\,{\scriptscriptstyle\pm}\,0.8$ & $\underline{72.4}^{(2)}\,{\scriptscriptstyle\pm}\,4.9$ & $21.5\,{\scriptscriptstyle\pm}\,1.0$ & $79.2^{(3)}\,{\scriptscriptstyle\pm}\,4.7$ & $95.0^{(3)}\,{\scriptscriptstyle\pm}\,1.7$ & $\mathbf{87.9}^{(1)}\,{\scriptscriptstyle\pm}\,7.2$\\
I + direct & $55.0\,{\scriptscriptstyle\pm}\,1.8$ & $40.4\,{\scriptscriptstyle\pm}\,2.8$ & $32.1\,{\scriptscriptstyle\pm}\,3.0$ & $39.8\,{\scriptscriptstyle\pm}\,1.7$ & $34.9\,{\scriptscriptstyle\pm}\,1.8$ & $54.2\,{\scriptscriptstyle\pm}\,3.1$ & $30.0\,{\scriptscriptstyle\pm}\,2.5$ & $23.4\,{\scriptscriptstyle\pm}\,1.3$ & $37.8\,{\scriptscriptstyle\pm}\,1.9$ & $59.1\,{\scriptscriptstyle\pm}\,2.7$ & $71.6\,{\scriptscriptstyle\pm}\,5.8$\\
I + H & $64.6\,{\scriptscriptstyle\pm}\,2.3$ & $54.1\,{\scriptscriptstyle\pm}\,5.0$ & $24.6\,{\scriptscriptstyle\pm}\,3.8$ & $45.0\,{\scriptscriptstyle\pm}\,2.9$ & $19.9\,{\scriptscriptstyle\pm}\,3.2$ & $78.5\,{\scriptscriptstyle\pm}\,1.1$ & $43.1\,{\scriptscriptstyle\pm}\,4.0$ & $5.8\,{\scriptscriptstyle\pm}\,0.6$ & $59.3\,{\scriptscriptstyle\pm}\,2.6$ & $86.2\,{\scriptscriptstyle\pm}\,4.3$ & $73.4\,{\scriptscriptstyle\pm}\,4.1$\\
S + direct & $68.3\,{\scriptscriptstyle\pm}\,2.1$ & $55.0\,{\scriptscriptstyle\pm}\,3.3$ & $41.2\,{\scriptscriptstyle\pm}\,5.6$ & $54.7\,{\scriptscriptstyle\pm}\,3.5$ & $51.8\,{\scriptscriptstyle\pm}\,2.1$ & $75.8\,{\scriptscriptstyle\pm}\,2.2$ & $53.5\,{\scriptscriptstyle\pm}\,6.3$ & $37.7\,{\scriptscriptstyle\pm}\,2.8$ & $59.6\,{\scriptscriptstyle\pm}\,4.1$ & $78.8\,{\scriptscriptstyle\pm}\,3.1$ & $80.6\,{\scriptscriptstyle\pm}\,7.9$\\
S + H & $\mathbf{74.8}^{(1)}\,{\scriptscriptstyle\pm}\,2.3$ & $69.3\,{\scriptscriptstyle\pm}\,5.4$ & $30.2\,{\scriptscriptstyle\pm}\,6.0$ & $57.1\,{\scriptscriptstyle\pm}\,8.0$ & $56.6\,{\scriptscriptstyle\pm}\,2.7$ & $88.0\,{\scriptscriptstyle\pm}\,0.5$ & $67.6\,{\scriptscriptstyle\pm}\,8.1$ & $15.5\,{\scriptscriptstyle\pm}\,1.4$ & $75.4\,{\scriptscriptstyle\pm}\,5.8$ & $94.6\,{\scriptscriptstyle\pm}\,3.3$ & $84.4\,{\scriptscriptstyle\pm}\,7.3$\\
RI + direct & $66.1\,{\scriptscriptstyle\pm}\,1.9$ & $57.9\,{\scriptscriptstyle\pm}\,4.4$ & $49.2\,{\scriptscriptstyle\pm}\,6.4$ & $53.8\,{\scriptscriptstyle\pm}\,4.0$ & $52.9\,{\scriptscriptstyle\pm}\,3.0$ & $76.7\,{\scriptscriptstyle\pm}\,2.6$ & $56.1\,{\scriptscriptstyle\pm}\,6.5$ & $41.4\,{\scriptscriptstyle\pm}\,2.6$ & $64.8\,{\scriptscriptstyle\pm}\,3.4$ & $81.0\,{\scriptscriptstyle\pm}\,2.8$ & $80.7\,{\scriptscriptstyle\pm}\,7.7$\\
RI + H & $71.7\,{\scriptscriptstyle\pm}\,1.8$ & $73.2^{(3)}\,{\scriptscriptstyle\pm}\,7.1$ & $48.1\,{\scriptscriptstyle\pm}\,11.1$ & $57.5\,{\scriptscriptstyle\pm}\,9.6$ & $63.5\,{\scriptscriptstyle\pm}\,3.6$ & $88.7^{(3)}\,{\scriptscriptstyle\pm}\,0.8$ & $70.1^{(3)}\,{\scriptscriptstyle\pm}\,5.7$ & $21.0\,{\scriptscriptstyle\pm}\,1.2$ & $78.6\,{\scriptscriptstyle\pm}\,4.4$ & $94.4\,{\scriptscriptstyle\pm}\,2.3$ & $86.5^{(3)}\,{\scriptscriptstyle\pm}\,8.5$\\
RS + direct & $68.9\,{\scriptscriptstyle\pm}\,2.1$ & $60.1\,{\scriptscriptstyle\pm}\,3.9$ & $50.9^{(3)}\,{\scriptscriptstyle\pm}\,6.7$ & $57.9\,{\scriptscriptstyle\pm}\,4.0$ & $58.5\,{\scriptscriptstyle\pm}\,2.3$ & $79.7\,{\scriptscriptstyle\pm}\,3.0$ & $59.5\,{\scriptscriptstyle\pm}\,6.9$ & $\mathbf{44.6}^{(1)}\,{\scriptscriptstyle\pm}\,3.0$ & $67.6\,{\scriptscriptstyle\pm}\,3.7$ & $84.1\,{\scriptscriptstyle\pm}\,2.9$ & $83.1\,{\scriptscriptstyle\pm}\,8.0$\\
RS + H & $74.3^{(3)}\,{\scriptscriptstyle\pm}\,1.9$ & $\underline{75.0}^{(2)}\,{\scriptscriptstyle\pm}\,5.1$ & $\mathbf{51.8}^{(1)}\,{\scriptscriptstyle\pm}\,12.4$ & $\mathbf{61.2}^{(1)}\,{\scriptscriptstyle\pm}\,8.9$ & $\underline{71.4}^{(2)}\,{\scriptscriptstyle\pm}\,3.2$ & $\mathbf{89.9}^{(1)}\,{\scriptscriptstyle\pm}\,0.6$ & $\mathbf{73.9}^{(1)}\,{\scriptscriptstyle\pm}\,4.5$ & $24.1\,{\scriptscriptstyle\pm}\,1.5$ & $\mathbf{80.8}^{(1)}\,{\scriptscriptstyle\pm}\,3.7$ & $\underline{95.7}^{(2)}\,{\scriptscriptstyle\pm}\,1.4$ & $\underline{87.2}^{(2)}\,{\scriptscriptstyle\pm}\,7.6$\\
IS + direct & $68.2\,{\scriptscriptstyle\pm}\,2.1$ & $56.9\,{\scriptscriptstyle\pm}\,3.7$ & $44.5\,{\scriptscriptstyle\pm}\,4.6$ & $54.1\,{\scriptscriptstyle\pm}\,2.4$ & $54.3\,{\scriptscriptstyle\pm}\,2.6$ & $75.9\,{\scriptscriptstyle\pm}\,1.8$ & $53.8\,{\scriptscriptstyle\pm}\,6.6$ & $38.4\,{\scriptscriptstyle\pm}\,2.8$ & $60.2\,{\scriptscriptstyle\pm}\,4.4$ & $79.1\,{\scriptscriptstyle\pm}\,3.2$ & $79.9\,{\scriptscriptstyle\pm}\,7.6$\\
IS + H & $\underline{74.6}^{(2)}\,{\scriptscriptstyle\pm}\,2.0$ & $71.9\,{\scriptscriptstyle\pm}\,5.3$ & $36.2\,{\scriptscriptstyle\pm}\,4.8$ & $57.9\,{\scriptscriptstyle\pm}\,8.2$ & $65.7^{(3)}\,{\scriptscriptstyle\pm}\,4.3$ & $87.0\,{\scriptscriptstyle\pm}\,0.4$ & $65.7\,{\scriptscriptstyle\pm}\,8.6$ & $16.2\,{\scriptscriptstyle\pm}\,1.5$ & $77.5\,{\scriptscriptstyle\pm}\,5.1$ & $94.3\,{\scriptscriptstyle\pm}\,2.2$ & $83.3\,{\scriptscriptstyle\pm}\,9.0$\\
RIS + direct & $68.9\,{\scriptscriptstyle\pm}\,2.1$ & $60.5\,{\scriptscriptstyle\pm}\,3.7$ & $50.3\,{\scriptscriptstyle\pm}\,7.3$ & $\underline{60.4}^{(2)}\,{\scriptscriptstyle\pm}\,7.5$ & $58.7\,{\scriptscriptstyle\pm}\,2.6$ & $79.9\,{\scriptscriptstyle\pm}\,2.4$ & $59.4\,{\scriptscriptstyle\pm}\,6.7$ & $\underline{44.4}^{(2)}\,{\scriptscriptstyle\pm}\,2.9$ & $67.6\,{\scriptscriptstyle\pm}\,3.9$ & $83.2\,{\scriptscriptstyle\pm}\,3.4$ & $82.4\,{\scriptscriptstyle\pm}\,8.3$\\
RIS + H & $74.0\,{\scriptscriptstyle\pm}\,1.9$ & $\mathbf{75.8}^{(1)}\,{\scriptscriptstyle\pm}\,4.9$ & $\underline{51.2}^{(2)}\,{\scriptscriptstyle\pm}\,12.2$ & $59.4^{(3)}\,{\scriptscriptstyle\pm}\,9.2$ & $\mathbf{72.0}^{(1)}\,{\scriptscriptstyle\pm}\,3.6$ & $\underline{89.6}^{(2)}\,{\scriptscriptstyle\pm}\,0.7$ & $68.8\,{\scriptscriptstyle\pm}\,7.4$ & $24.0\,{\scriptscriptstyle\pm}\,1.6$ & $\underline{80.4}^{(2)}\,{\scriptscriptstyle\pm}\,3.5$ & $\mathbf{95.8}^{(1)}\,{\scriptscriptstyle\pm}\,1.4$ & $85.8\,{\scriptscriptstyle\pm}\,8.5$\\
\bottomrule\end{tabularx}
\end{table}
\end{landscape}
\clearpage

\subsection{Representation and decoding effects}

Table~\ref{tab:effects_tests} reports paired representation and decoding effects across all four metrics.

\begin{table}[H]\centering\footnotesize\setlength{\tabcolsep}{4pt}\renewcommand{\arraystretch}{0.90}
\caption{Controlled effects over nine primary dataset means, in percentage points with 95\% dataset-bootstrap intervals. For $R/I/S$, additions average six contexts and full-context contrasts add the representation to the other two, averaging the two decoding modes; $H$ effects average seven representations. W/T/L counts positive/zero/negative dataset effects. Exact sign-flip $p$-values are Holm-adjusted separately within each metric's three addition tests and three full-context tests; the single decoding test is unchanged.}\label{tab:effects_tests}
\begin{tabularx}{\linewidth}{@{}Xlrrrc@{}}
\toprule Measure & Contrast & Change & 95\% interval & $p_H$ & W/T/L\\\midrule
Macro-F1 & add R & +10.24 & [+7.69, +12.68] & 0.012 & 9/0/0\\
 & add I & +0.49 & [-0.25, +1.26] & 0.273 & 6/0/3\\
 & add S & +8.04 & [+6.46, +10.32] & 0.012 & 9/0/0\\
 & full R & +4.80 & [+3.12, +6.62] & 0.012 & 9/0/0\\
 & full I & -0.27 & [-0.93, +0.20] & 0.555 & 3/0/6\\
 & full S & +3.01 & [+2.09, +4.20] & 0.012 & 9/0/0\\
 & H vs direct & +5.39 & [-2.39, +11.65] & 0.199 & 7/0/2\\
\addlinespace[3pt]
Accuracy & add R & +8.56 & [+6.30, +10.71] & 0.012 & 9/0/0\\
 & add I & +0.58 & [+0.16, +1.05] & 0.031 & 8/0/1\\
 & add S & +7.08 & [+5.53, +9.29] & 0.012 & 9/0/0\\
 & full R & +3.79 & [+2.24, +5.54] & 0.012 & 9/0/0\\
 & full I & -0.04 & [-0.27, +0.18] & 0.770 & 3/0/6\\
 & full S & +2.63 & [+1.91, +3.81] & 0.012 & 9/0/0\\
 & H vs direct & +6.35 & [-0.85, +12.95] & 0.129 & 6/0/3\\
\addlinespace[3pt]
Balanced accuracy & add R & +8.66 & [+5.91, +11.37] & 0.012 & 9/0/0\\
 & add I & +0.21 & [-0.61, +1.04] & 0.703 & 4/0/5\\
 & add S & +7.19 & [+5.56, +9.46] & 0.012 & 9/0/0\\
 & full R & +3.81 & [+2.03, +5.87] & 0.016 & 8/0/1\\
 & full I & -0.43 & [-0.95, +0.01] & 0.145 & 2/0/7\\
 & full S & +2.73 & [+1.95, +3.90] & 0.012 & 9/0/0\\
 & H vs direct & -0.16 & [-9.18, +8.07] & 0.992 & 6/0/3\\
\addlinespace[3pt]
Worst-class F1 & add R & +5.08 & [+1.63, +9.05] & 0.070 & 7/1/1\\
 & add I & +0.11 & [-0.75, +1.33] & 0.953 & 3/1/5\\
 & add S & +4.75 & [+2.01, +7.60] & 0.070 & 6/1/2\\
 & full R & +2.36 & [-0.74, +5.61] & 0.406 & 5/2/2\\
 & full I & -0.04 & [-0.52, +0.49] & 0.906 & 3/2/4\\
 & full S & +2.58 & [+1.04, +4.19] & 0.094 & 6/2/1\\
 & H vs direct & +2.27 & [-4.06, +9.56] & 0.602 & 3/1/5\\
\addlinespace[3pt]
\bottomrule\end{tabularx}\end{table}

\end{document}